\pdfoutput=1

\documentclass[11pt]{article}

\usepackage[final]{acl}

\usepackage{times}
\usepackage{latexsym}

\usepackage[T1]{fontenc}

\usepackage[utf8]{inputenc}

\usepackage{microtype}

\usepackage{inconsolata}

\usepackage{graphicx}

\usepackage{xcolor}
\usepackage[linesnumbered,ruled,vlined]{algorithm2e}
\usepackage{CJKutf8}
\usepackage{multirow}
\usepackage{graphicx}
\usepackage{subcaption}
\usepackage{float}
\usepackage{enumitem}
\usepackage{pifont}
\usepackage{booktabs}

\title{Sequential-NIAH: A Needle-In-A-Haystack Benchmark for Extracting Sequential \emph{Needles} from Long Contexts}

\author{Yifei Yu, Qian-Wen Zhang\textsuperscript{$\dagger$}, Lingfeng Qiao, Di Yin, Fang Li, Jie Wang,\\
\textbf{Zengxi Chen, Suncong Zheng, Xiaolong Liang, Xing Sun} \\
Tencent Youtu Lab, China \\
cowenzhang@tencent.com, felixyfyu@tencent.com
}

\begin{document}
\maketitle
\def\thefootnote{$\dagger$}\footnotetext{Corresponding Author.}
\begin{abstract}
Evaluating the ability of large language models (LLMs) to process lengthy contexts is critical, especially for retrieving query-relevant information embedded within them. We introduce Sequential-NIAH, a benchmark specifically designed to evaluate the capability of LLMs to extract sequential information items (known as \emph{needles}) from long contexts. The benchmark includes three needle generation pipelines: synthetic-temporal, real-temporal, and real-logical orders, with context lengths ranging from 8K to 128K, which comprises 14,000 samples (2,000 for testing). To facilitate the evaluation of this benchmark, we trained an evaluation model that assesses the correctness of LLM responses by comparing their completeness and sequential consistency against the ground truth, which provides a more reliable evaluation metric than GPT-4 or Claude. We conducted experiments on six well-known LLMs, revealing that even the best-performing model achieved a maximum accuracy of only 63.50\% on test set of this benchmark. Further analysis highlights the growing challenges posed by increasing the context length or the number of needles, underscoring substantial room for improvement of LLMs. Additionally, noise analysis validates the reliability and challenge of the benchmark, making Sequential-NIAH an important reference for advancing research on long text information extraction capabilities of LLMs.
\end{abstract}

\section{Introduction}

\begin{figure*}[t]
  \includegraphics[width=\textwidth]{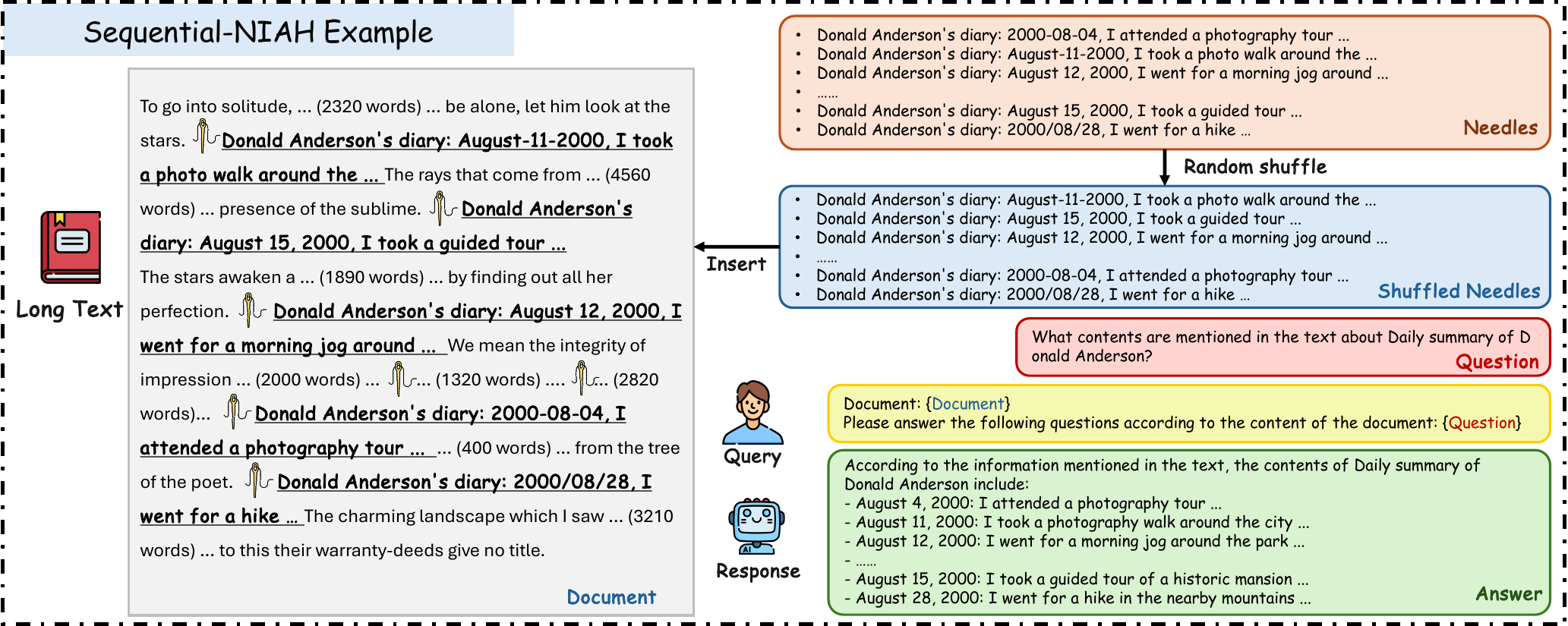}
  \caption{Sequential-NIAH example of a long text with shuffled needles with temporal order.}
  \label{fig:example}
\end{figure*}

Enhancing LLMs' long-context understanding has been a key focus in Natural Language Processing (NLP). Recent models like Gemini-1.5 \cite{gemini}, GPT-4 \cite{achiam2023gpt}, Claude-3.5 \cite{claude}, Qwen-2.5 \cite{qwen2.5}, GLM-4 \cite{GLM-4}, Kimi \cite{kimi}, and DeepSeek-V2 \cite{deepseek} have extended context lengths to millions of tokens while maintaining reasoning and comprehension capabilities. Meanwhile, several benchmarks have been exposed for long context understanding, including $\infty$Bench \cite{infinitebench}, L-Eval \cite{leval}, LongBench \cite{longbench}, LongEval \cite{longeval}, LooGLE \cite{loogle}, ZeroSCROLLS \cite{zeroscrolls} and FactGuard \cite{factguard}. However, these benchmarks typically focus more on the model's global comprehension of long texts or the retrieval of specific information at certain locations. In reality, challenging problems often involve retrieving and integrating detailed information from multiple parts of a document to produce the optimal answer, which can generally be defined as \emph{Needle-in-a-Haystack} (NIAH) tasks \cite{NIAH, ruler, countingstar, needlebench}.

Although existing NIAH benchmarks provide challenging needle-retrieval tasks within long contexts, they still fail to account for the sequential characteristic of \emph{needles}—such as temporal or logical order. This oversight is particularly significant in real-world scenarios, where explicit demands exist, such as: 
\begin{itemize}
    \item \emph{List all events involving suspect Tom in March 2024 in temporal order, based on a legal document.}
    \item \emph{List all Microsoft equity transactions in temporal order, based on a financial report.}
    \item \emph{Outline the detailed steps to obtain a senior building engineer certification in sequential order, as per a guideline.}
\end{itemize}
Figure~\ref{fig:example} provides a simple example illustrating the sequential relationships among needles, simulating a more realistic and challenging NIAH task. For LLMs, it is essential to not only retrieve query-relevant items but also comprehend their sequential relationships and present them in the correct order. 

To supplement existing long context information retrieval evaluation methods, We introduce the Sequential-NIAH benchmark, which shuffles \emph{needles} with temporal or logical order and inserts them into long contexts of varying lengths. Considering the completeness of the benchmark, we propose three different types of \emph{needles} generation pipelines, including synthetic-temporal order, real-temporal order and real-logical order. Synthetic-temporal order \emph{needles} are generated by fake entities, timestamps, and events. Real-temporal order \emph{needles} are generated from the Temporal Knowledge Graph (TKG\cite{icews14,icews18,FEG}), which can be used to build sequential temporal items based on the relationship of two entities overtime. Real-logical order \emph{needles} are generated from a private open-domain QA resource. The first two types are mainly aimed at retrieving temporal sequence items, while the last type is mainly aimed at retrieving logical sequence items.

On the other hand, accurately evaluating the correctness of enumerated answer items is challenging to achieve. The common practice relies on assessment by powerful LLMs, such as GPT-4o, which increases evaluation costs and hinders researchers from conducting efficient benchmarking. Thus, we employed synthetic data to train an evaluation model. The synthetic dataset encompasses as diverse a range of incorrect answer types as possible (e.g., missing items, redundancies, errors, disordered items, etc.). These will be paired with reference answers as training data to teach the evaluation model to accurately identify incorrect responses. The validation results indicate that the evaluation model achieved an accuracy of 99.49\% (much higher than GPT-4o and Claude) on the synthetic test set, which is a reliable evaluation tool for our benchmark.

\begin{figure*}[t]
  \includegraphics[width=\textwidth]{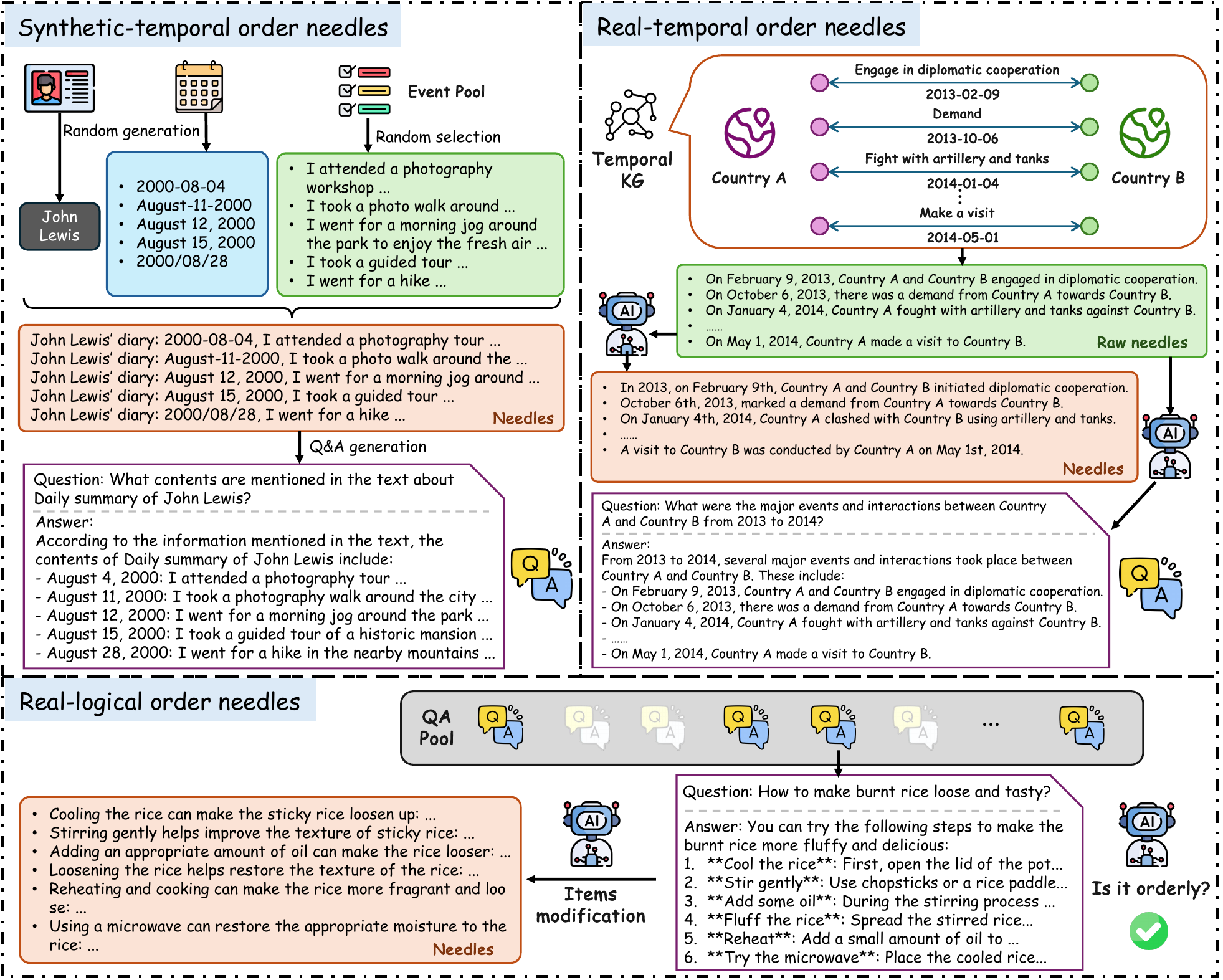}
  \caption{Three pipelines for sequential needles construction. Synthetic temporal order needles are generated by fake subjects, time stamps, and events (upper left). Real temporal orders are generated from the TKG (upper right). Real logical order needles are generated from a private open-domain QA resource (lower).}
  \label{fig:PPL}
\end{figure*}

Powered by our evaluation model, we assessed the accuracy of several popular LLMs on the benchmark. Results show the task is highly challenging, with the best model reaching only 63.50\% accuracy. Longer contexts and more needles further increase difficulty of the task. A noise robustness analysis also confirmed the benchmark’s reliability and challenge. Our key contributions are:
\begin{itemize} 
\item \textbf{Sequential-NIAH benchmark}: A benchmark for evaluating LLMs' ability to retrieve sequential information from long contexts. It comprises three types of \emph{needles} generation pipelines, covering both temporal-ordered and logical-ordered \emph{needles} retrieval tasks, to simulate real-world application scenarios.
\item \textbf{Evaluation model}: A model trained on synthetic data to facilitate the evaluation of LLMs' performance on our benchmark. An accuracy rate of 99.49\% demonstrates the model's high reliability as an evaluation tool, which is more accurate, efficient and cheaper than GPT-4o and Claude.
\item \textbf{LLMs' Limitations on Sequential-NIAH}: Experimental results indicate that all current LLMs have significant room for improvement on this task, struggling with the complexity of sequential information retrieval within long contexts.
\end{itemize}

\section{Related work}

\subsection{Long Context Language Models}

\begin{table*}[t]
  \centering
  \resizebox{\textwidth}{!}{
  \begin{tabular}{lrrrr|lrrrr}
  \hline
  \multicolumn{5}{c|}{QA source} & \multicolumn{5}{c}{Long context source} \\ \hline
  Pipeline & English & Chinese & Total & Proportion & Length & English & Chinese & Total & Proportion \\ \hline
  Synthetic-temporal & 3,000 & 3,000 & 6,000 & 42.86\% & 8k-16k & 1,000 & 750 & 1,750 & 12.50\% \\
  Real-temporal & 3,003 & 2,011 & 5,014 & 35.81\% & 16k-32k & 2,000 & 1,750 & 3,750 & 26.79\% \\ 
  Real-logical & 1,497 & 1,489 & 2,986 & 21.33\% & 32k-64k & 2,000 & 1,750 & 3,750 & 26.79\% \\
  Total & 7,500 & 6,500 & 14,000 & & 64k-128k & 2,500 & 2,250 & 4,750 & 33.92\% \\
  Proportion & 46.43\% & 53.57\% & & & Total & 7,500 & 6,500 & 14,000 \\ 
  \hline
  \end{tabular}}
  \caption{Information of QA source from three sequential needles synthetic pipelines (left) and long context source extracted from LongData-Corpus (right).}
  \label{tab:QAinfo}
\end{table*}

Many techniques have been used to improve the context length that LLMs can handle. For instance, certain novel position embedding methods, such as ALiBi \cite{alibi}, Position Interpolation \cite{PI}, RoPE \cite{RoPE} and its variants \cite{RoPE_1, RoPE_2, RoPE_3}. And some research aims to reduce context length by memory replay back-propagation \cite{memformer}, recurrent memory augmentation \cite{RMT}, and activation beacon \cite{AB}. In addition, there are several methods to extend the context length by modifying the model architecture, such as Mamba \cite{mamba}, FLASHBUTTERFLY \cite{HELC}, and RWKV \cite{RWKV}. Specifically, Gemini-1.5 supports a context length of 1 million tokens, and Kimi supports a context length of 2 million words. Most leading LLMs support a context length of at least 128k tokens, such as GPT-4o, Claude-3.5, Qwen-2.5, Qwen-3, and LLaMA-3.3. In this work, we will evaluate these LLMs with long-context analysis capabilities to assess their performance on our benchmark.

\subsection{NIAH Benchmarks and Tasks}

NIAH essentially represents a category of long-text information retrieval tasks, primarily assessing the capability of LLMs to retrieve multiple pieces of query-relevant information from long texts. The RULER \cite{ruler} and Counting Starts \cite{countingstar} benchmarks are designed with retrieval tasks at the word or character level (passwords or \ding{72}), where the problems involved are relatively clear and singular. NeedleBench \cite{needlebench} takes this one step further by designing more complex information on logical reasoning, such as descriptions of relationships between entities or kinship relationships, and inserting them into a long context. In Sequential-NIAH, we designed a challenging long-context information extraction task by needles with sequential characteristics, to better reflect real-world use cases.

\subsection{Evaluation Model}

The evaluation of natural language generation (NLG) is
a vital but challenging problem in artificial intelligence \cite{evaluation_survey}. Its primary methods include the following four types: LLM derived metrics \cite{gptscore, zs-F}, prompt-based LLMs \cite{instruct, luo2303chatgpt, gao2023humanlikesummarizationevaluationchatgpt, liusie2024llmcomparativeassessmentzeroshot}, fine-tuning LLMs \cite{instructscore, CritiqueLLM, JudgeLM}, and collaborative human-LLM evaluation \cite{zhang-etal-2021-human, li2023collaborativeevaluationexploringsynergy}. Commonly, simply designing prompts often fails to achieve optimal accuracy in evaluation, and fine-tuning on open-source models can enhance the accuracy of the evaluation model effectively. If manpower is sufficient, the combination of human effort with LLMs can further improve the reliability of the evaluation. To facilitate a reliable automatic evaluation of Sequential-NIAH tasks, we trained an evaluation model based on Qwen2.5-Instruct-32B, which provides a convenient and reliable evaluation tool for this benchmark.

\section{The Sequential-NIAH Benchmark}

The goal of Sequential-NIAH is to retrieve sequential needles from long contexts. Therefore, both the needles and the long contexts need to be prepared in advance. See the link\footnote{\url{https://github.com/miraclefish/Sequential-NIAH-Benchmark.git}} for details of the dataset.

\subsection{QA source}

We propose three sequential needles generation pipelines, as shown in Figure~\ref{fig:PPL}, which are used to build the question with sequential answer items, including synthetic-temporal order needles, real-temporal order needles and real-logical order needles. All are ultimately presented in the form of a question with multiple answer items (\emph{needles}) inserted into a long text with length from 8K to 128K. Table~\ref{tab:QAinfo} provides detailed information of the number and proportion of QA pairs constructed by different pipelines, collectively referred to as \textbf{QA source}. We adjusted the proportion of Chinese and English QA pairs to maintain each around 50\%.

\subsubsection{Synthetic-temporal order needles}

Synthetic-temporal order needles refers to the synthesis of question-answer pairs using specific generation templates by combining subjects, event times, and event content. The question is usually posed about events that occur within a certain time period for a predefined fake subject, and the answer items are the synthetic events related to the fake subject listed in temporal order. In theory, this method can generate an unlimited number of qualified chronological question-answer pairs. We ensure the complexity of the task by designing various question templates and needle templates. The number of needles corresponding to a question can be set manually, and we randomly select the number of needles from 3 to 15.

\subsubsection{Real-temporal order needles}

From open source TKG datasets (ICEWS and FEG), we can extract real relationships between two different real entities change over time, which can be used to generate real-temporal order needles. In our pipeline, the relationships are rewritten (by GPT-4o) into question-answer pairs with temporal order answer items. The question is usually posed about the changes in relationships between two specific entities over a period of time. The amount of data that can be generated by this pipeline is limited by the size of the Graph, and the number of needles corresponding to a question depends on the number of relationships between two entities (from 3 to 10 in this pipeline).

\begin{algorithm}[t]
    \SetAlgoLined
    \KwData{long text $C$, question $Q$, answer $A$, needles $N=[n_1, n_2, \dots, n_k]$.}
    \KwResult{$Query$ and $Answer$.}

    $[C_1, C_2, \dots, C_{k+1}]\leftarrow$ Segment($C,k+1$)\;
    $N\leftarrow$ Shuffle($N$)\;

    Long text with needles: $\hat{C}\leftarrow C_1$\;
    \For{$i \leftarrow 1$ \KwTo $k$}{
        $\hat{C} \leftarrow \hat{C} + n_i + C_{i+1}$\;
    }
    $Query \leftarrow$ Prompt($\hat{C}, Q$)\;
    $Answer \leftarrow A$\;  
    \Return $Query$ and $Answer$\;
    \caption{Data construction pipeline}
    \label{alg:samplePPL}
\end{algorithm}

\subsubsection{Real-logical order needles}

In addition to items in temporal order, there are also cases where answers follow a precise logical order. To incorporate these into the benchmark, we filtered out question-answer pairs that meet the requirements from a private open-domain QA database with the help of GPT-4o. A total of 2,986 QA pairs are collected, whose answer items strictly adhere to a logical order. We manually conducted a sampling check on the filtered data to ensure the reliability of the QA filtering. Considering that directly inserting the answer items into the long context might seem abrupt (making it difficult to establish a direct connection between the question and the needles), we also use GPT-4o to rewrite the answer items to generate more naturally phrased needles. This ensures that when needles are inserted into the long context, they can still make connections to the question, maintaining the rationality of the task.

\subsection{Long Context Source}

To enhance the authenticity of the task, we use LongData-Corpus \cite{LongDataCorpus}, a real long text corpus, to construct the \textbf{long context source}. The corpus contains more than 100k pieces of Chinese and English long texts with lengths exceeding 8k characters, with the longest text exceeding 256k characters. The text content covers a wide range of materials such as academic papers, novels, legal documents, news, patents, government work reports, etc. This provides ample long context data for the construction of the benchmark. When preparing the long context source, to keep the language and quantity of long texts consistent with the QA source, we randomly sample long texts within different length ranges for each language (Chinese and English), as shown in Table~\ref{tab:QAinfo}. This forms a long context source that covers a wide enough range of topics, has a reasonable distribution of article lengths, and can match each QA pair in the QA source one-to-one.

\begin{table}[t]
  \centering
  \begin{tabular}{lrrr}
  \hline
  & English & Chinese & Total \\ \hline
  Train & 5,400  & 4,600 & 10,000 \\
  Development & 1,015 & 985 & 2,000 \\
  Test & 1085 & 915 & 2,000 \\
  Total & 7,500 & 6,500 & 14,000 \\
  \hline
  \end{tabular}
  \caption{Dataset information of Sequential-NIAH Benchmark}
  \label{tab:BenchmarkInfo}
\end{table}

\subsection{Sequential-NIAH Sample Constructing}

For a given long text and a QA pair, a specific Sequential-NIAH sample is constructed by randomly shuffling the order of needles in the answer of QA and inserting them into random positions within the long text. Subsequently, the sample is formatted into $Query$ (inserted long text and question) and $Answer$ (reference answer) forms using a designed prompt template, as shown in Figure~\ref{fig:example}. The detailed procedure is described in Algorithm~\ref{alg:samplePPL}. Ultimately, the dataset is partitioned into three subsets: training set, development set, and test set. Each subset contains data from diverse languages, varying text lengths, and distinct needles synthetic pipelines, embodying a Sequential-NIAH benchmark, as shown in Table~\ref{tab:BenchmarkInfo}.

\begin{table}[b]
  \centering
  \begin{tabular}{lrrr}
  \hline
  & English & Chinese & Total \\ 
  \hline
  Train & 3,000  & 3,000 & 6,000 \\
  Test & 984 & 967 & 1,960 \\
  Total & 3,984 & 3,967 & 7,960 \\
  \hline
  \end{tabular}
  \caption{Dataset information for evaluation model training and test.}
  \label{tab:em_dataset}
\end{table}


\section{Evaluation Model}

\begin{figure*}[t]
  \includegraphics[width=\textwidth]{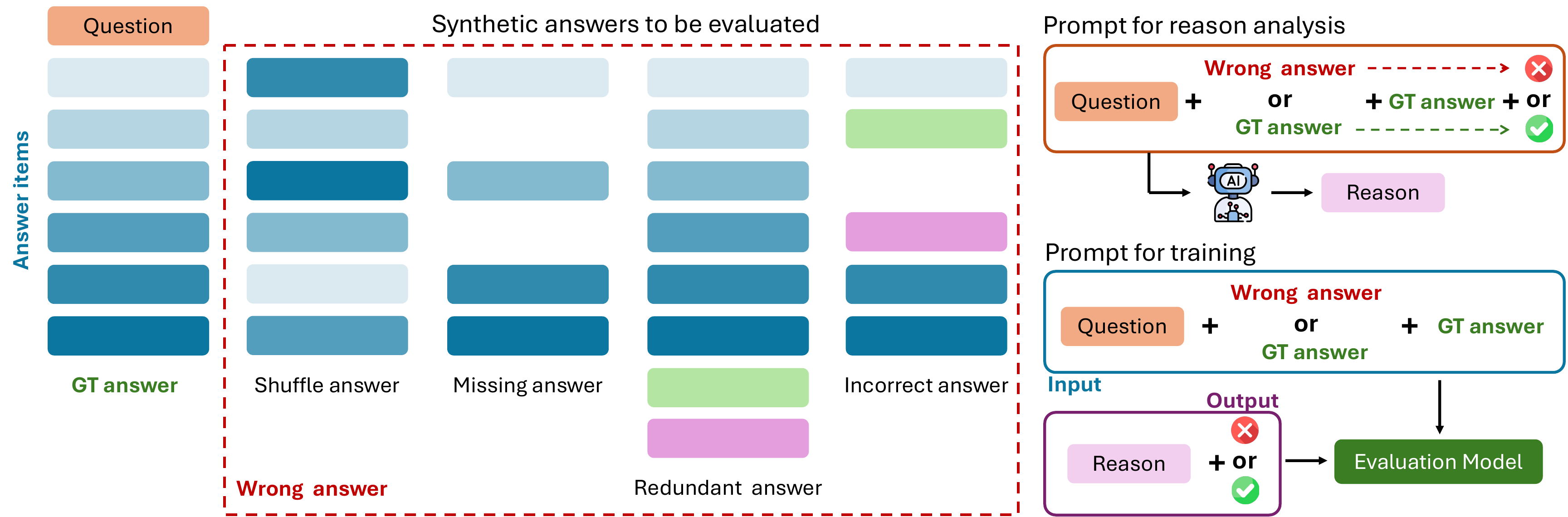}
  \caption{Synthetic different wrong answers for generating diversified training data for evaluation model training.}
  \label{fig:EM}
\end{figure*}

\begin{table}[t]
  \centering
  \footnotesize
  \renewcommand{\arraystretch}{1.2} 
  \resizebox{\columnwidth}{!}{
  \begin{tabular}{p{0.8cm}p{1cm}rp{1.2cm}p{1cm}p{0.5cm}}
    \hline
        Needle types & Answer groups & No. & Claude-3.5 (\%) & GPT-4o (\%) & Ours (\%) \\ \hline
        \multirow{4}{0.8cm}{Temporal order} & GT & 241 & 85.06 & 90.87 & \textbf{95.85} \\
        & Missing & 235 & 93.62 & 97.45 & \textbf{100} \\ 
        & Redundant & 265 & 77.74 & 85.28 & \textbf{100} \\ 
        & Incorrect & 229 & 95.20 & 97.38 & \textbf{100} \\ 
        \hline
        \multirow{4}{0.8cm}{Logical order} & GT & 229 & 86.03 & 99.13 & \textbf{100 }\\ 
        & Missing & 249 & 93.17 & \textbf{100} & \textbf{100} \\
        & Redundant & 266 & 85.34 & 99.25 & \textbf{100} \\ 
        & Incorrect & 246 & 82.11 & \textbf{100} & \textbf{100} \\ 
        \hline
        & Total/Avg. & 1960 & 87.09 & 96.07 & \textbf{99.49} \\
    \hline
    \end{tabular}}
  \caption{Evaluation model performance on synthetic test data with various needle types and answer groups.}
  \label{tab:em_test_result}
\end{table}

Due to the complexity of benchmark evaluation, we hope to automate the evaluation of this task by training an evaluation model $f_{\theta}$. For each question $Q_i$, the ground truth answer $A_i$ and an corresponding answer $B_i$ to be evaluated are provided to constitute an input $X_i=T_{eval}(Q_i,A_i,B_i)$, where $T_{eval}(\cdot)$ is a prompt template for answer evaluation. For each $X_i$, the label $Y_i=T_{res}(y_i,R_i)$ is constructed by the result ($y_i\in\{wrong,correct\}$) and the reason $R_i$, where $T_{res}(\cdot)$ is the prompt template for result analysis. To train the evaluation model, our objective is to learn the probability distribution of $Y_i$ conditioned on $X_i$, i.e., $P(Y_i|X_i;\theta)$. And the loss function can be defined as:

\begin{equation}
    \mathcal{L}(\theta) = \arg\max_{\theta} \sum_{i=1}^{N} \log P(Y_i | X_i; \theta)
\end{equation}

To obtain the training data, four types of potential wrong answers are synthesized by changing GT answer items, including shuffled answer items (shuffled GT answer items), missing answer items (GT answer items with random missing items), redundant answer items (GT answer items and random redundant items), and incorrect answer items (random missing items and random redundant items coexist). For the last three groups of answers, we uniformly consider them as wrong answers. For the first group of answer with only shuffled items, if the question does not require the answer items to be output in specific order, it will be treated as correct answers; otherwise, it will be treated as an wrong answers.

Figure~\ref{fig:EM} provides an example of how to generate training data for evaluation model training. At first, four kinds of wrong answer are generated according to the ground truth (GT) answer. For each question $Q_i$ we can get an answer pair by combining the GT answer $A_i$ and the answer $B_i$ to be evaluated. Then, $Q_i$, $A_i$, $B_i$ and the known result $y_i$ are organized into the reason analysis prompt template to get the reason by GPT-4o. Finally, the obtained reason $R_i$ and $y_i$ can be combined as the output $Y_i$, and $Q_i$, $A_i$ and $B_i$ will be organized as the input $X_i$ for evaluation model training.

As shown in Table~\ref{tab:em_dataset}, a total of 6,000 samples are constructed to train the evaluation model, and 1,960 samples are used to evaluate its performance. The data used to train and test the evaluation model are randomly sampled from the QA source. We used Qwen2.5-Instruct-32B as the foundation for our evaluation model and performed full-parameter SFT training on it. We utilized the AdamW optimizer, setting the learning rate to $8\times 10^{-6}$ with 4 epoch. We set the warm-up ratio to 0.1 and the weight decay to 0.1.

\section{Experiments \& Results}

\subsection{Evaluation Model Performance}

To demonstrate the need to train the evaluation model, we compared the performance of Claude-3.5, GPT-4o, and our evaluation model on 1,960 test samples using the same prompt templates ($T_{eval}$ and $T_{res}$). The experimental results are shown in Table~\ref{tab:em_test_result}, and the results are divided into two groups (needles with temporal and logical order) for analysis. More detailed grouped results are presented in Table~\ref{tab:em_test_detail}.

According to analysis, the evaluation model we trained achieved a total accuracy rate of \textbf{99.49\%} in two groups of test data, which is much higher than GPT-4o and Claude-3.5. Our evaluation model achieved an accuracy of 100\% in 7 answer groups. And the only 0.51\% misjudgment of our evaluation model came from the model's slight confusion about whether the question requires listing answer items in temporal order (only occurring in shuffled GT answer items). This sufficiently demonstrates that using the model for automated evaluation of the benchmark is entirely feasible.

\begin{figure}[t]
  \includegraphics[width=\linewidth]{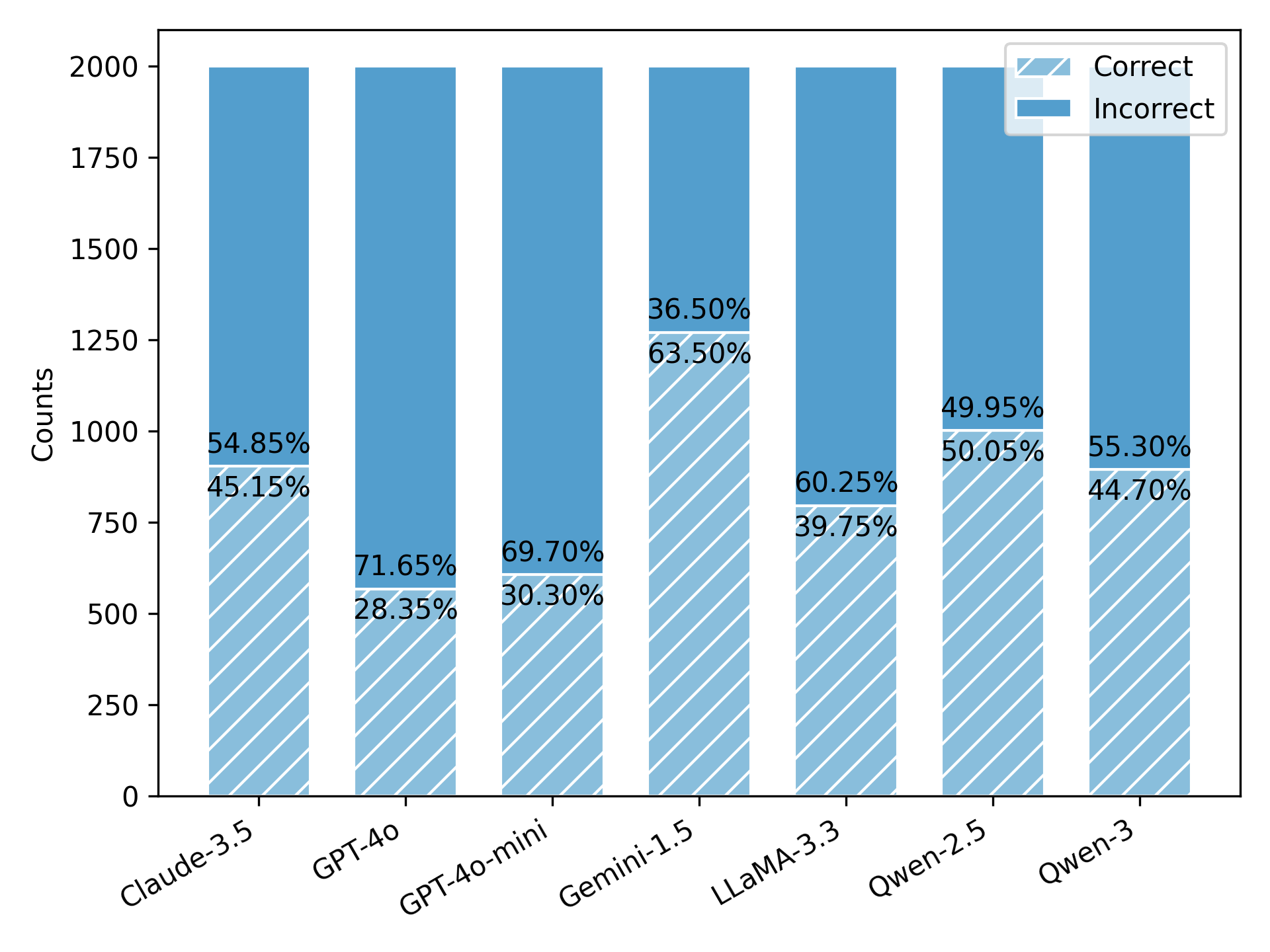}
  \caption{Results comparison of LLMs on all test data of Sequential-NIAH benchmark.}
  \label{fig:all_test_result}
\end{figure}

\subsection{Benchmark Results of LLMs}

To evaluate the performance of different LLMs on this benchmark, we conducted inference on 2,000 test samples using four closed-source models, including \textbf{Claude-3.5} (Claude-3.5-sonnet-20241022), \textbf{GPT-4o} (GPT-4o-20240806), \textbf{GPT-4o-mini}, \textbf{Gemini-1.5} (Gemini-1.5-pro), and three open-source models, including \textbf{Qwen-2.5} (Qwen2.5-72B-Instruct), \textbf{Qwen-3} (Qwen3-32B), and \textbf{LLaMA-3.3} (LLaMA-3.3-70B-Instruct).

\begin{figure}[t]
  \includegraphics[width=\linewidth]{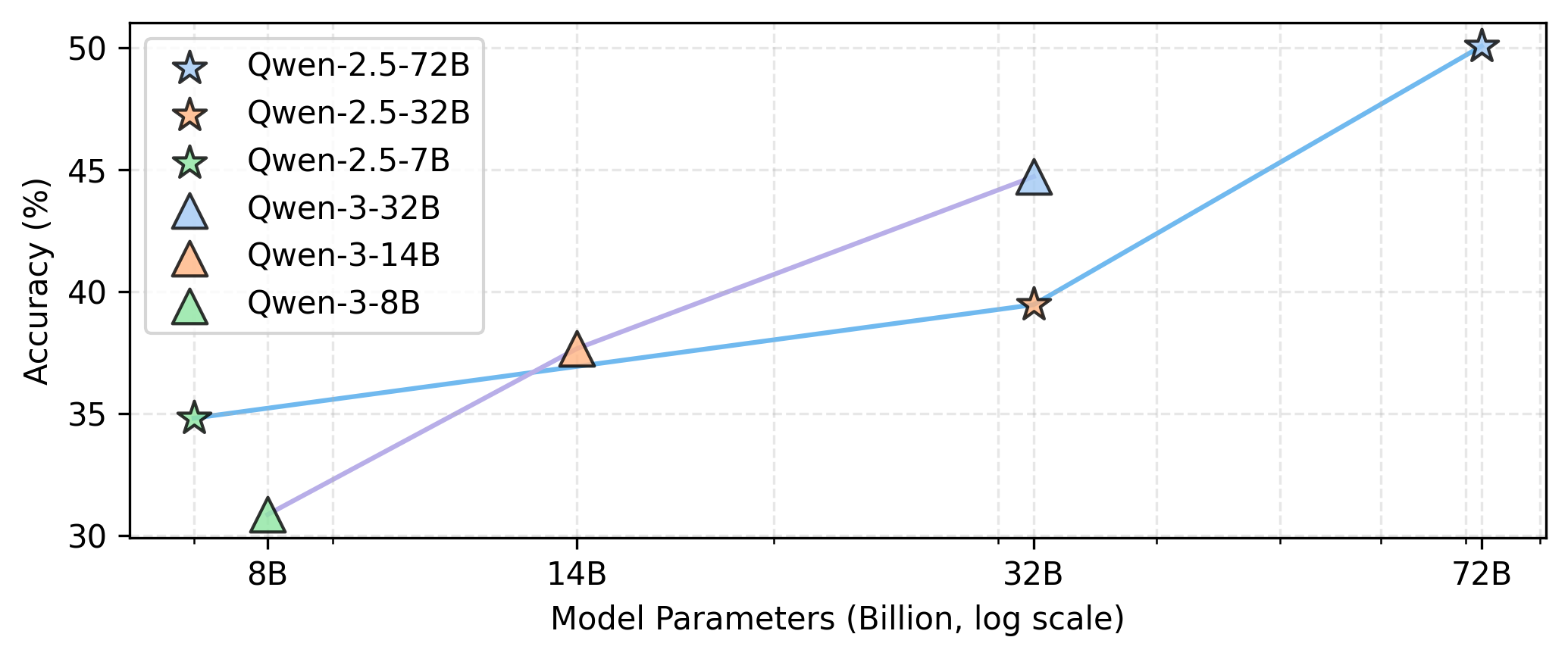}
  \caption{Performance of model size scaling on Qwen-2.5 and Qwen-3 series.}
  \label{fig:scaling_up_result}
\end{figure}

\textbf{Overview of results}: Figure~\ref{fig:all_test_result} illustrates the overall performance of LLMs on this benchmark. Gemini-1.5 exhibits the best performance, achieving an accuracy of 63.50\%. Qwen-2.5 follows closely behind with an accuracy of 50.05\%, while LLaMA-3.3, Qwen-3 and Claude-3.5 demonstrate comparable levels of performance. In contrast, GPT-4o-mini and GPT-4o perform poorly on this task. Moreover, experiments on Qwen-2.5 and Qwen-3 with size scaling show that performance degrades as the model size decreases, as shown in Figure~\ref{fig:scaling_up_result}.

\begin{figure*}[ht]
    \centering
    \begin{subfigure}[b]{0.48\textwidth}
        \includegraphics[width=\textwidth]{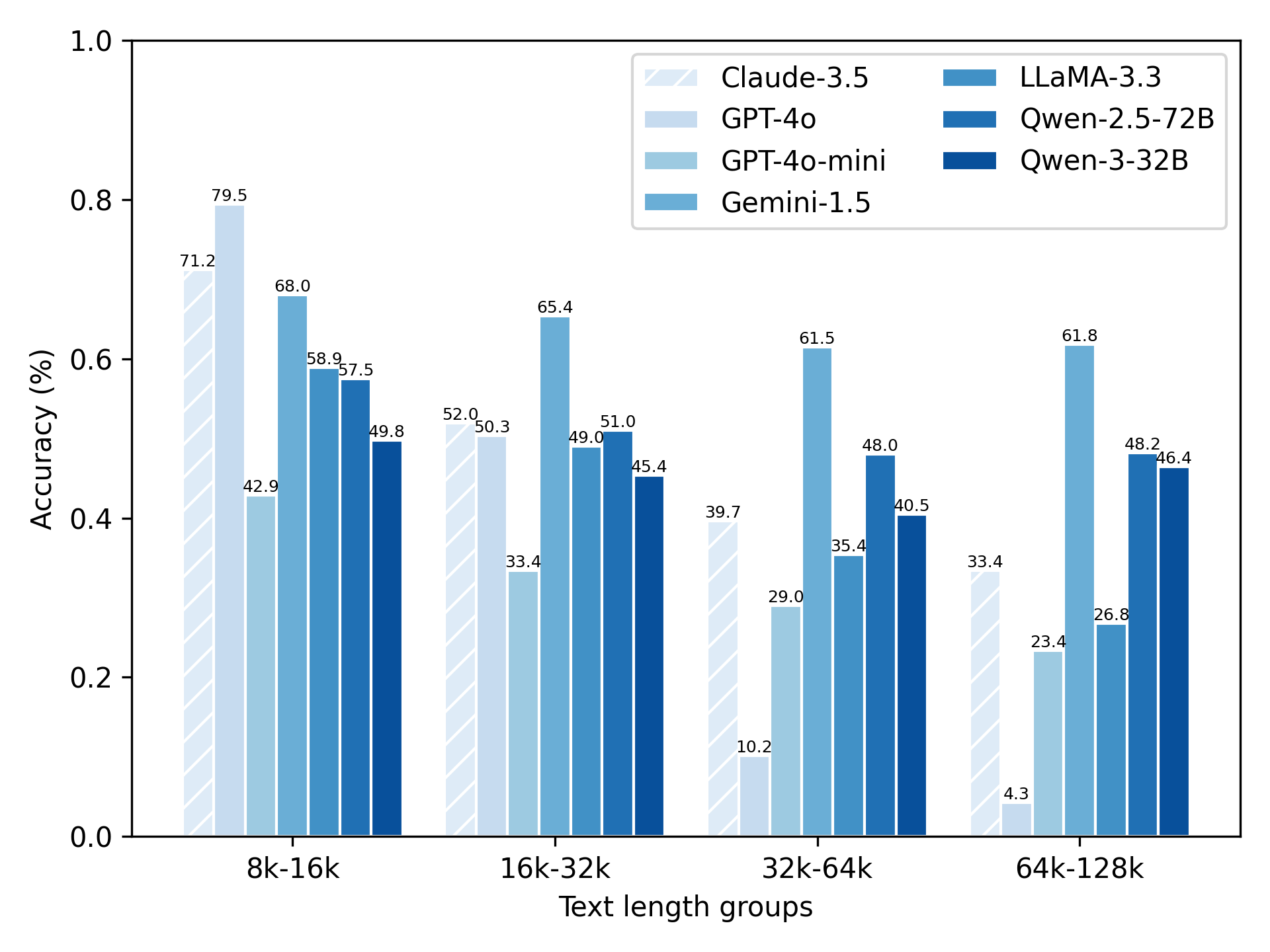}
        \caption{Results across length groups.}
        \label{fig:result_1}
    \end{subfigure}
    \hfill
    \begin{subfigure}[b]{0.48\textwidth}
        \includegraphics[width=\textwidth]{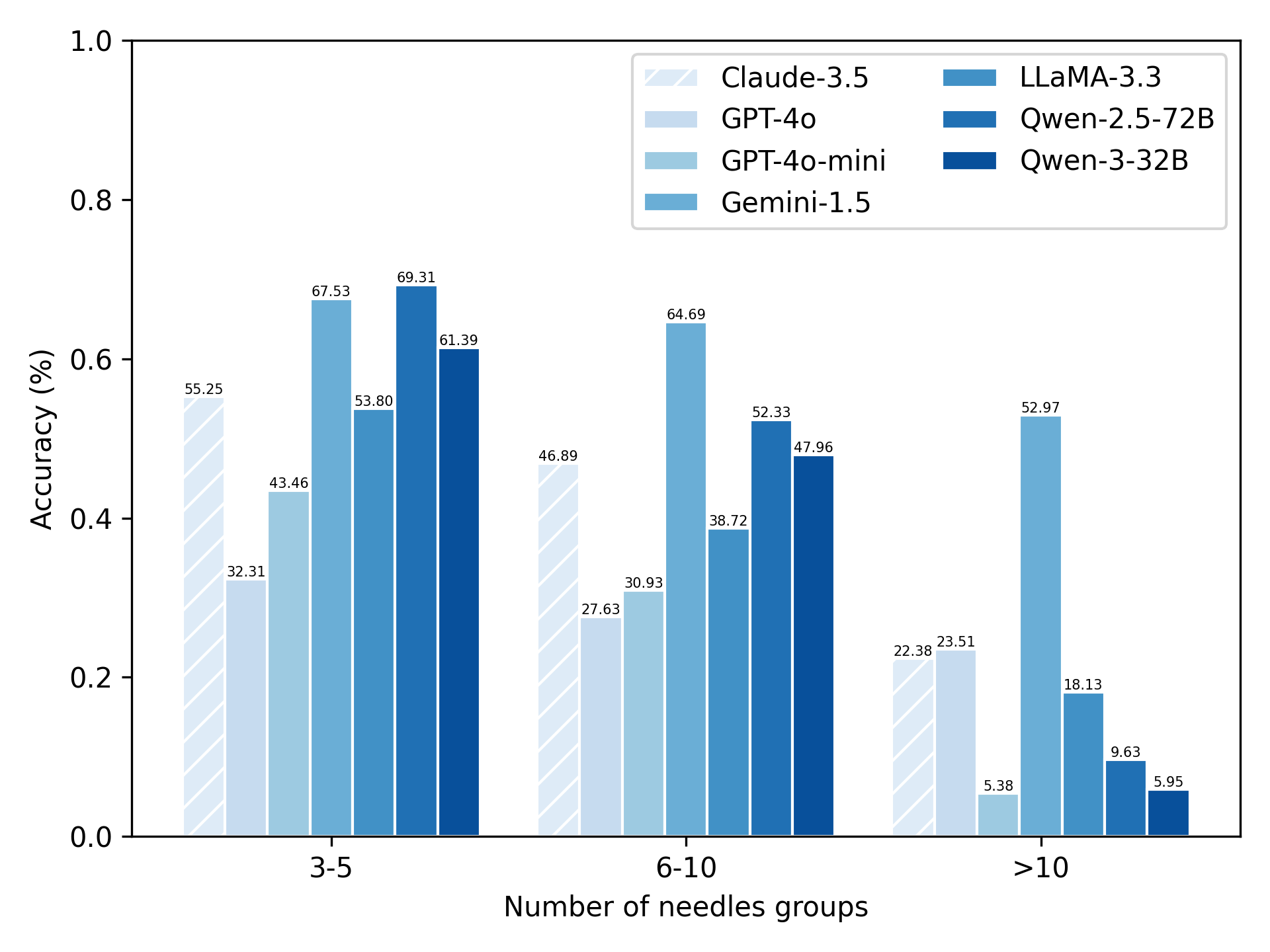}
        \caption{Results across the number of needles groups.}
        \label{fig:result_2}
    \end{subfigure}
    \begin{subfigure}[b]{0.48\textwidth}
        \includegraphics[width=\textwidth]{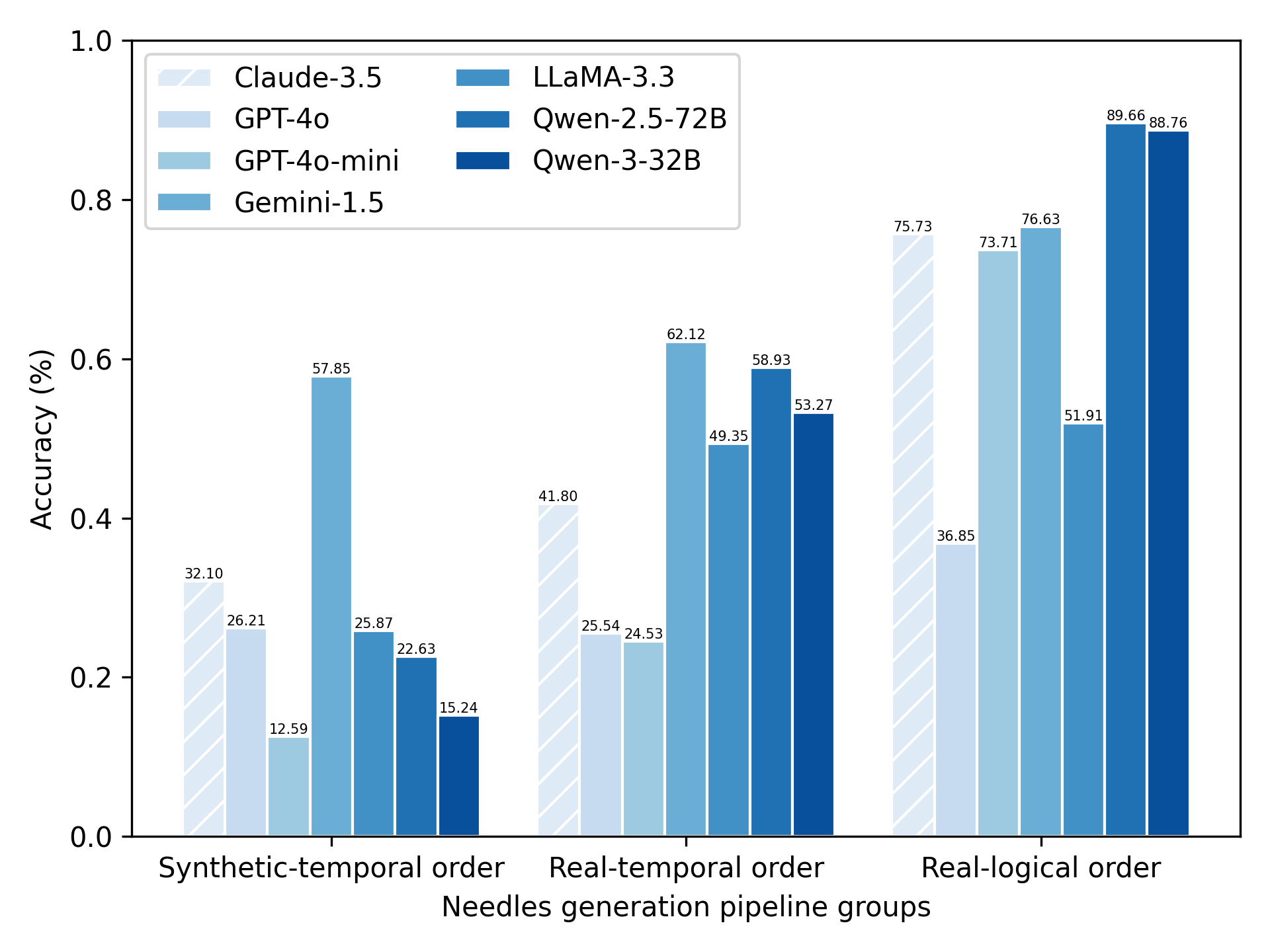}
        \caption{Results across needles generation pipeline groups.}
        \label{fig:result_3}
    \end{subfigure}
    \hfill
    \begin{subfigure}[b]{0.48\textwidth}
        \includegraphics[width=\textwidth]{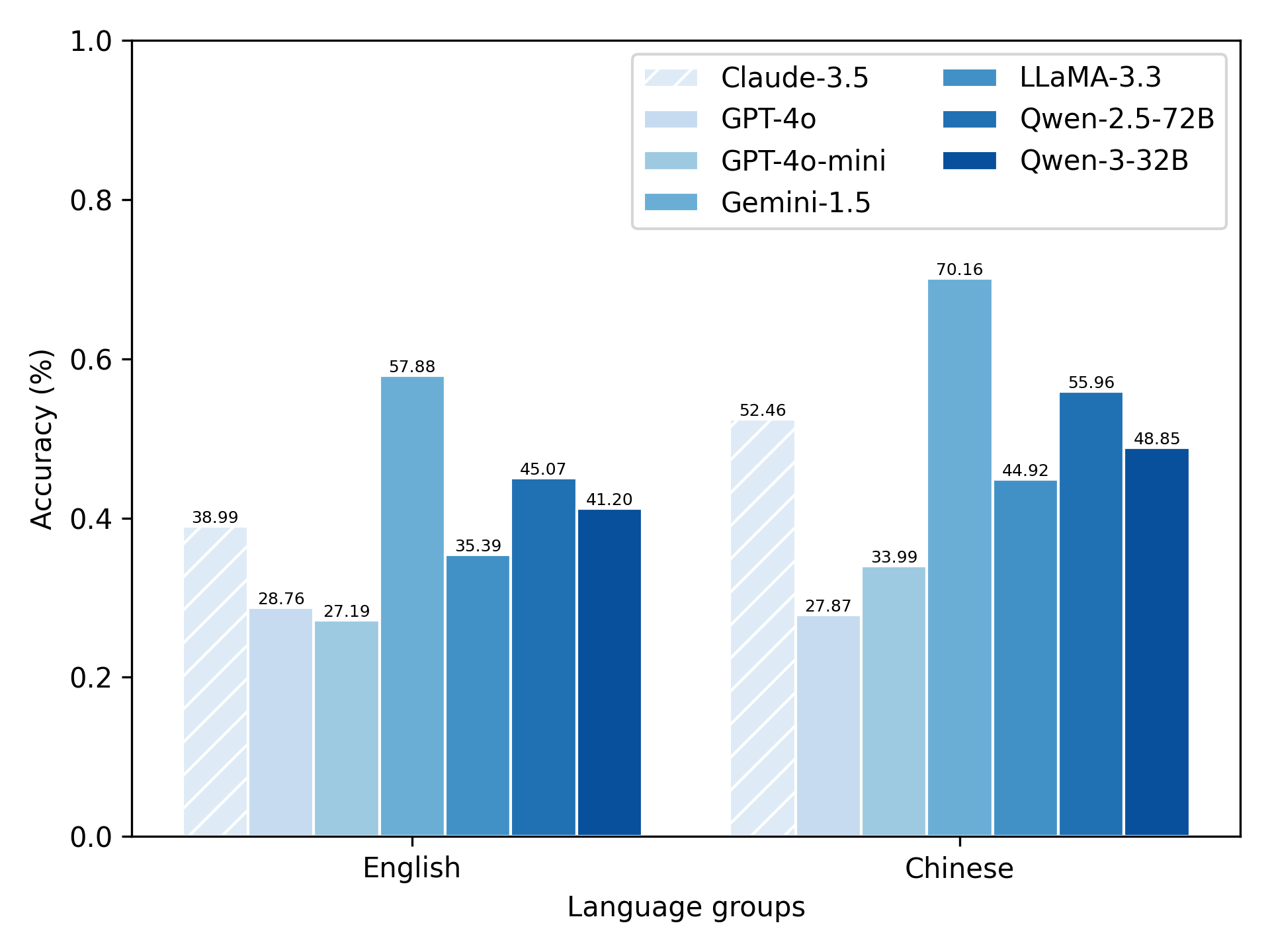}
        \caption{Results across language groups.}
        \label{fig:result_4}
    \end{subfigure}
    \caption{Benchmark results of well-known LLMs on test data across different groups.}
    \label{fig:combined}
\end{figure*}

\textbf{Results across length groups}: Figure~\subref{fig:result_1} shows that the accuracy of most LLMs decreases as the text length increases. However, Gemini-1.5 and Qwen-2.5 maintain better and more stable performance. LLaMA-3.3 and GPT-4o decrease significantly with longer text.

\textbf{Results across the number of needles groups}: Intuitively, more needles will significantly increase the difficulty of this task. Figure~\subref{fig:result_2} shows that all LLMs effects deteriorate as the number of needles increases. Surprisingly, Gemini-1.5 still maintains more stable accuracy compared with others.

\textbf{Results across needles generation pipeline groups}: Figure~\subref{fig:result_3} presents that most LLMs perform better on test data composed of real-logical order needles. It may be attributed to the fact that the questions in this group include some general knowledge, allowing LLMs to provide answers close to the GT based on their inherent capabilities, rather than retrieving from long texts. It also indicates that retrieving and listing information in temporal order from long texts is more challenging.

\textbf{Results across language groups}: Figure~\subref{fig:result_4} depicts that the performance of the same model on test samples in different languages is generally consistent, indicating that language is not a key factor affecting the difficulty of the task.

\subsection{Noise Analysis of the Benchmark}

In our investigation of this benchmark's characteristics, we selected 200 samples from the test set to conduct a noise analysis. Noise analysis, in this context, refers to evaluating the stability of various LLMs' performance when the needles or the long context face variations that may affect response. Specifically, we introduced four distinct types of noise to each sample:
\begin{itemize}
    \item \textbf{Tiny movement (TM)}: Each needle within the long text undergoes a slight positional shift, either forward or backward, by no more than two sentence positions. The order of the needles within the long text remains unchanged
    \item \textbf{Significant movement (SM)}: Multiple needles within the long text are repositioned significantly, with their original sequence maintained. This simulates larger displacements, enabling analysis of the model's robustness to more pronounced positional alterations.
    \item \textbf{Reorder (RO)}: The positions of multiple needles remain static, but their sequence of appearance within the text is shuffled.
    \item \textbf{Semantic noise (SN)}: One or two synthetic needles that are semantically similar to one needle—yet cannot serve as a valid answer item—are inserted at somewhere in the long context, simulating scenarios that may confuse the model's judgment.
\end{itemize}
For each type of noise, we generated three variations of the original data, culminating in a total of 2400 noisy samples used for inference and evaluation. In this section, three specific models, \textbf{Gemini-1.5}, \textbf{Qwen-2.5}, and \textbf{LLaMA-3.3}, are subjected to the noise analysis experiments to discern their resilience and performance under these controlled perturbations.

\begin{table}[t]
    \centering
    \footnotesize
    \renewcommand{\arraystretch}{1.2}
    \resizebox{\linewidth}{!}{
    \begin{tabular}{lp{0.8cm}p{1.2cm}p{0.5cm}p{0.5cm}p{0.5cm}p{0.5cm}p{0.5cm}}
    \hline
        Model & Metrics & Ref. (\%) & TM (\%) & SM (\%) & RO (\%) & SN (\%) & All (\%)  \\ \hline
        \multirow{3}{0.8cm}{Gemini-1.5} & \multirow{2}{0.6cm}{Acc.} & \multirow{2}{0.6cm}{62.50} & 65.00 & 63.12 & 64.12 & 53.88 & 61.31  \\
        &&& $\pm$2.55 & $\pm$1.89 & $\pm$1.80 & $\pm$7.36 & $\pm$6.58 \\
        \cline{2-8}
        & Cons. & - & 56.50 & 52.00 & 47.50 & 41.50 & 17.50  \\ \hline
        \multirow{3}{0.8cm}{Qwen-2.5} & \multirow{2}{0.6cm}{Acc.} & \multirow{2}{0.6cm}{51.50} & 48.50 & 48.25 & 48.00 & 43.00 & 45.88  \\
        &&& $\pm$2.00 & $\pm$2.25 & $\pm$2.68 & $\pm$6.10 & $\pm$3.73 \\
        \cline{2-8}
        & Cons. & - & 67.50 & 65.50 & 64.50 & 64.50 & 42.00  \\ \hline
        \multirow{3}{0.8cm}{LLaMA-3.3} & \multirow{2}{0.6cm}{Acc.} & \multirow{2}{0.6cm}{38.00} & 38.00 & 36.50 & 42.00 & 29.88 & 36.27  \\
        &&& $\pm$0.71 & $\pm$2.38 & $\pm$2.92 & $\pm$5.78 & $\pm$6.07 \\
        \cline{2-8}
        & Cons. & - & 63.50 & 55.50 & 55.50 & 62.00 & 29.50 \\ \hline
    \end{tabular}}
    \caption{Average accuracy and result consistency of LLMs with different noise assigned on test data. `All' indicates that all noise groups are collectively included in the metric calculation.}
    \label{tab:noise}
\end{table}

Two metrics are employed for the noise analysis experiment: average accuracy (\textbf{Acc.}) and result consistency (\textbf{Cons.}). \textbf{Acc.} represents the mean accuracy across the original 200 test samples and additional test samples with introduced noise. \textbf{Cons.} assesses the stability of the model's responses by comparing the consistency of answers' evaluation result of the original 200 test samples with those of the noise-altered sets.

\textbf{Average accuracy analysis}: In Table~\ref{tab:noise}, the reference accuracy (\textbf{Ref.}) represents the original accuracy achieved by the 200 samples drawn from the test set. Under full-noise conditions (column `All'), the Acc. of the three models deviates from the reference values by 1.19\% for Gemini-1.5, 5.62\% for Qwen-2.5 and 1.73\% for LLaMA-3.3. While exhibiting model-specific variance, all LLMs demonstrate quantitatively acceptable performance deviations (no more than 6\%). It demonstrates that the test set exhibits consistent reliability in evaluating LLMs' ability on this benchmark, with evaluation results remaining robust against both minor and major needles variations.

\textbf{Result Consistent analysis}: Table~\ref{tab:noise} also shows that the LLMs exhibit varying degrees of response stability under noise influence, with Qwen performing best, LLaMA second, and Gemini worst. This may occur because LLaMA's responses contain a relatively high proportion of incorrect answers, and the introduction of noise fails to correct these errors, resulting in consistently higher error rates in its outputs. On the other hand, the more noise groups are introduced, the worse the Cons. becomes, which clearly demonstrates that noise can significantly impact the correctness of LLM responses, further highlighting the challenging of this benchmark.

\section{Conclusion}

We introduce Sequential-NIAH, a benchmark for evaluating LLMs on sequential information extraction from long texts (up to 128K tokens). It includes synthetic-temporal, real-temporal, and real-logical order needles generation pipelines, with 10K/2K/2K train/dev/test splits, and an evaluation model for efficient assessment.  

Experiments show Claude, GPT-4o, Gemini, LLaMA, and Qwen struggle with the benchmark, revealing its complexity and the need for model improvements. Noise analysis confirms its reliable and challenging, marking a valuable contribution to the NLP community.

\section*{Limitations}

Model evaluations may be biased by the dataset's domain, and unoptimized API parameters could affect performance and fairness. Addressing these is crucial for accurate assessments.

The dataset is for academic and research use only; commercial or misuse is prohibited to maintain integrity and ethical standards.

\bibliography{custom}

\begin{thebibliography}{46}
\providecommand{\natexlab}[1]{#1}

\bibitem[{Achiam et~al.(2024)Achiam, Adler, Agarwal, Ahmad, Akkaya, Aleman, Almeida, Altenschmidt, Altman, Anadkat et~al.}]{achiam2023gpt}
Josh Achiam, Steven Adler, Sandhini Agarwal, Lama Ahmad, Ilge Akkaya, Florencia~Leoni Aleman, Diogo Almeida, Janko Altenschmidt, Sam Altman, Shyamal Anadkat, and 1 others. 2024.
\newblock \href {https://arxiv.org/abs/2303.08774} {Gpt-4 technical report}.
\newblock \emph{arXiv preprint arXiv:2303.08774}.

\bibitem[{An et~al.(2023)An, Gong, Zhong, Zhao, Li, Zhang, Kong, and Qiu}]{leval}
Chenxin An, Shansan Gong, Ming Zhong, Xingjian Zhao, Mukai Li, Jun Zhang, Lingpeng Kong, and Xipeng Qiu. 2023.
\newblock \href {https://arxiv.org/abs/2307.11088} {L-eval: Instituting standardized evaluation for long context language models}.
\newblock \emph{arXiv preprint arXiv:2307.11088}.

\bibitem[{Anthropic(2024)}]{claude}
Anthropic. 2024.
\newblock Introducing the next generation of claude 3.5.
\newblock \url{https://www.anthropic.com/news/claude-3-5-sonnet}.

\bibitem[{Bai et~al.(2024)Bai, Lv, Zhang, Lyu, Tang, Huang, Du, Liu, Zeng, Hou, Dong, Tang, and Li}]{longbench}
Yushi Bai, Xin Lv, Jiajie Zhang, Hongchang Lyu, Jiankai Tang, Zhidian Huang, Zhengxiao Du, Xiao Liu, Aohan Zeng, Lei Hou, Yuxiao Dong, Jie Tang, and Juanzi Li. 2024.
\newblock \href {https://arxiv.org/abs/2308.14508} {Longbench: A bilingual, multitask benchmark for long context understanding}.
\newblock \emph{arXiv preprint arXiv:2308.14508}.

\bibitem[{Bulatov et~al.(2024)Bulatov, Kuratov, Kapushev, and Burtsev}]{RMT}
Aydar Bulatov, Yuri Kuratov, Yermek Kapushev, and Mikhail~S. Burtsev. 2024.
\newblock \href {https://arxiv.org/abs/2304.11062} {Scaling transformer to 1m tokens and beyond with rmt}.
\newblock \emph{Preprint}, arXiv:2304.11062.

\bibitem[{Chen et~al.(2023)Chen, Wong, Chen, and Tian}]{PI}
Shouyuan Chen, Sherman Wong, Liangjian Chen, and Yuandong Tian. 2023.
\newblock \href {https://doi.org/10.48550/arXiv.2306.15595} {Extending {{Context Window}} of {{Large Language Models}} via {{Positional Interpolation}}}.
\newblock \emph{Preprint}, arXiv:2306.15595.

\bibitem[{Fu et~al.(2023{\natexlab{a}})Fu, Epstein, Nguyen, Thomas, Zhang, Dao, Rudra, and R{\'e}}]{HELC}
Daniel~Y Fu, Elliot~L Epstein, Eric Nguyen, Armin~W Thomas, Michael Zhang, Tri Dao, Atri Rudra, and Christopher R{\'e}. 2023{\natexlab{a}}.
\newblock Simple hardware-efficient long convolutions for sequence modeling.
\newblock In \emph{International Conference on Machine Learning}, pages 10373--10391. PMLR.

\bibitem[{Fu et~al.(2023{\natexlab{b}})Fu, Ng, Jiang, and Liu}]{gptscore}
Jinlan Fu, See-Kiong Ng, Zhengbao Jiang, and Pengfei Liu. 2023{\natexlab{b}}.
\newblock \href {https://arxiv.org/abs/2302.04166} {Gptscore: Evaluate as you desire}.
\newblock \emph{Preprint}, arXiv:2302.04166.

\bibitem[{Gao et~al.(2024)Gao, Hu, Ruan, Pu, and Wan}]{evaluation_survey}
Mingqi Gao, Xinyu Hu, Jie Ruan, Xiao Pu, and Xiaojun Wan. 2024.
\newblock \href {https://arxiv.org/abs/2402.01383} {Llm-based nlg evaluation: Current status and challenges}.
\newblock \emph{Preprint}, arXiv:2402.01383.

\bibitem[{Gao et~al.(2023)Gao, Ruan, Sun, Yin, Yang, and Wan}]{gao2023humanlikesummarizationevaluationchatgpt}
Mingqi Gao, Jie Ruan, Renliang Sun, Xunjian Yin, Shiping Yang, and Xiaojun Wan. 2023.
\newblock \href {https://arxiv.org/abs/2304.02554} {Human-like summarization evaluation with chatgpt}.
\newblock \emph{Preprint}, arXiv:2304.02554.

\bibitem[{Garc{\'\i}a-Dur{\'a}n et~al.(2018)Garc{\'\i}a-Dur{\'a}n, Duman{\v{c}}i{\'c}, and Niepert}]{icews14}
Alberto Garc{\'\i}a-Dur{\'a}n, Sebastijan Duman{\v{c}}i{\'c}, and Mathias Niepert. 2018.
\newblock \href {https://arxiv.org/abs/1809.03202} {Learning sequence encoders for temporal knowledge graph completion}.
\newblock \emph{arXiv preprint arXiv:1809.03202}.

\bibitem[{Georgiev et~al.(2024)Georgiev, Lei, Burnell, Bai, Gulati, Tanzer, Vincent, Pan, Wang et~al.}]{gemini}
Petko Georgiev, Ving~Ian Lei, Ryan Burnell, Libin Bai, Anmol Gulati, Garrett Tanzer, Damien Vincent, Zhufeng Pan, Shibo Wang, and 1 others. 2024.
\newblock \href {https://arxiv.org/abs/2403.05530} {Gemini 1.5: Unlocking multimodal understanding across millions of tokens of context}.
\newblock \emph{arXiv preprint arXiv:2403.05530}.

\bibitem[{gkamradt(2023)}]{NIAH}
gkamradt. 2023.
\newblock \href {https://github.com/gkamradt/LLMTest_NeedleInAHaystack} {Needle in a haystack - pressure testing llms}.

\bibitem[{Gu and Dao(2024)}]{mamba}
Albert Gu and Tri Dao. 2024.
\newblock \href {https://arxiv.org/abs/2312.00752} {Mamba: Linear-time sequence modeling with selective state spaces}.
\newblock \emph{Preprint}, arXiv:2312.00752.

\bibitem[{Hsieh et~al.(2024)Hsieh, Sun, Kriman, Acharya, Rekesh, Jia, Zhang, and Ginsburg}]{ruler}
Cheng-Ping Hsieh, Simeng Sun, Samuel Kriman, Shantanu Acharya, Dima Rekesh, Fei Jia, Yang Zhang, and Boris Ginsburg. 2024.
\newblock \href {https://arxiv.org/abs/2404.06654} {Ruler: What's the real context size of your long-context language models?}
\newblock \emph{arXiv preprint arXiv:2404.06654}.

\bibitem[{Jia et~al.(2023)Jia, Ren, Liu, and Zhu}]{zs-F}
Qi~Jia, Siyu Ren, Yizhu Liu, and Kenny~Q. Zhu. 2023.
\newblock \href {https://arxiv.org/abs/2310.11648} {Zero-shot faithfulness evaluation for text summarization with foundation language model}.
\newblock \emph{Preprint}, arXiv:2310.11648.

\bibitem[{Jin et~al.(2020)Jin, Qu, Jin, and Ren}]{icews18}
Woojeong Jin, Meng Qu, Xisen Jin, and Xiang Ren. 2020.
\newblock \href {https://arxiv.org/abs/1904.05530} {Recurrent event network: Autoregressive structure inference over temporal knowledge graphs}.
\newblock \emph{arXiv preprint arXiv:1904.05530}.

\bibitem[{Ke et~al.(2024)Ke, Wen, Feng, Liu, Lei, Cheng, Wang, Zeng, Dong, Wang, Tang, and Huang}]{CritiqueLLM}
Pei Ke, Bosi Wen, Zhuoer Feng, Xiao Liu, Xuanyu Lei, Jiale Cheng, Shengyuan Wang, Aohan Zeng, Yuxiao Dong, Hongning Wang, Jie Tang, and Minlie Huang. 2024.
\newblock \href {https://arxiv.org/abs/2311.18702} {Critiquellm: Towards an informative critique generation model for evaluation of large language model generation}.
\newblock \emph{Preprint}, arXiv:2311.18702.

\bibitem[{Krishna et~al.(2023)Krishna, Bransom, Kuehl, Iyyer, Dasigi, Cohan, and Lo}]{longeval}
Kalpesh Krishna, Erin Bransom, Bailey Kuehl, Mohit Iyyer, Pradeep Dasigi, Arman Cohan, and Kyle Lo. 2023.
\newblock \href {https://arxiv.org/abs/2301.13298} {Longeval: Guidelines for human evaluation of faithfulness in long-form summarization}.
\newblock \emph{arXiv preprint arXiv:2301.13298}.

\bibitem[{Li et~al.(2024{\natexlab{a}})Li, Wang, Zheng, and Zhang}]{loogle}
Jiaqi Li, Mengmeng Wang, Zilong Zheng, and Muhan Zhang. 2024{\natexlab{a}}.
\newblock \href {https://arxiv.org/abs/2311.04939} {Loogle: Can long-context language models understand long contexts?}
\newblock \emph{arXiv preprint arXiv:2311.04939}, arXiv:2311.04939.

\bibitem[{Li et~al.(2024{\natexlab{b}})Li, Zhang, Liu, and Chen}]{needlebench}
Mo~Li, Songyang Zhang, Yunxin Liu, and Kai Chen. 2024{\natexlab{b}}.
\newblock \href {https://arxiv.org/abs/2407.11963} {Needlebench: Can llms do retrieval and reasoning in 1 million context window?}
\newblock \emph{arXiv preprint arXiv:2407.11963}.

\bibitem[{Li et~al.(2023)Li, Cui, Kong, and Bi}]{li2023collaborativeevaluationexploringsynergy}
Qintong Li, Leyang Cui, Lingpeng Kong, and Wei Bi. 2023.
\newblock \href {https://arxiv.org/abs/2310.19740} {Collaborative evaluation: Exploring the synergy of large language models and humans for open-ended generation evaluation}.
\newblock \emph{Preprint}, arXiv:2310.19740.

\bibitem[{Liu et~al.(2024{\natexlab{a}})Liu, Feng, Wang, Wang, Liu, Zhao, Dengr, Ruan, Dai, Guo et~al.}]{deepseek}
Aixin Liu, Bei Feng, Bin Wang, Bingxuan Wang, Bo~Liu, Chenggang Zhao, Chengqi Dengr, Chong Ruan, Damai Dai, Daya Guo, and 1 others. 2024{\natexlab{a}}.
\newblock \href {https://arxiv.org/abs/2405.04434} {Deepseek-v2: A strong, economical, and efficient mixture-of-experts language model}.
\newblock \emph{arXiv preprint arXiv:2405.04434}.

\bibitem[{Liu et~al.(2024{\natexlab{b}})Liu, Bai, Zhang, Zhang, Zhang, Zhang, Wang, Que, Chen, Su, Ge, Fu, Chen, and Zheng}]{RoPE_2}
Jiaheng Liu, Zhiqi Bai, Yuanxing Zhang, Chenchen Zhang, Yu~Zhang, Ge~Zhang, Jiakai Wang, Haoran Que, Yukang Chen, Wenbo Su, Tiezheng Ge, Jie Fu, Wenhu Chen, and Bo~Zheng. 2024{\natexlab{b}}.
\newblock \href {https://arxiv.org/abs/2401.06951} {E$^2$llm: Efficient and extreme length extension of large language models}.
\newblock \emph{Preprint}, arXiv:2401.06951.

\bibitem[{liuhuanyong(2022)}]{FEG}
liuhuanyong. 2022.
\newblock \href {https://github.com/liuhuanyong/FinanceEventGraph} {Finance event graph}.

\bibitem[{Liusie et~al.(2024)Liusie, Manakul, and Gales}]{liusie2024llmcomparativeassessmentzeroshot}
Adian Liusie, Potsawee Manakul, and Mark J.~F. Gales. 2024.
\newblock \href {https://arxiv.org/abs/2307.07889} {Llm comparative assessment: Zero-shot nlg evaluation through pairwise comparisons using large language models}.
\newblock \emph{Preprint}, arXiv:2307.07889.

\bibitem[{Luo et~al.(2023)Luo, Xie, and Ananiadou}]{luo2303chatgpt}
Z~Luo, Q~Xie, and S~Ananiadou. 2023.
\newblock Chatgpt as a factual inconsistency evaluator for abstractive text summarization (2023).
\newblock \emph{arXiv preprint arXiv:2303.15621}.

\bibitem[{Moonshot()}]{kimi}
Moonshot.
\newblock Kimi.
\newblock \url{https://www.moonshot.cn}.

\bibitem[{Ouyang et~al.(2022)Ouyang, Wu, Jiang, Almeida, Wainwright, Mishkin, Zhang, Agarwal, Slama, Ray, Schulman, Hilton, Kelton, Miller, Simens, Askell, Welinder, Christiano, Leike, and Lowe}]{instruct}
Long Ouyang, Jeffrey Wu, Xu~Jiang, Diogo Almeida, Carroll Wainwright, Pamela Mishkin, Chong Zhang, Sandhini Agarwal, Katarina Slama, Alex Ray, John Schulman, Jacob Hilton, Fraser Kelton, Luke Miller, Maddie Simens, Amanda Askell, Peter Welinder, Paul~F Christiano, Jan Leike, and Ryan Lowe. 2022.
\newblock \href {https://proceedings.neurips.cc/paper_files/paper/2022/file/b1efde53be364a73914f58805a001731-Paper-Conference.pdf} {Training language models to follow instructions with human feedback}.
\newblock In \emph{Advances in Neural Information Processing Systems}, volume~35, pages 27730--27744. Curran Associates, Inc.

\bibitem[{Peng et~al.(2023{\natexlab{a}})Peng, Alcaide, Anthony, Albalak, Arcadinho, Biderman, Cao, Cheng, Chung, Grella, GV, He, Hou, Lin, Kazienko, Kocon, Kong, Koptyra, Lau, Mantri, Mom, Saito, Song, Tang, Wang, Wind, Wozniak, Zhang, Zhang, Zhao, Zhou, Zhou, Zhu, and Zhu}]{RWKV}
Bo~Peng, Eric Alcaide, Quentin Anthony, Alon Albalak, Samuel Arcadinho, Stella Biderman, Huanqi Cao, Xin Cheng, Michael Chung, Matteo Grella, Kranthi~Kiran GV, Xuzheng He, Haowen Hou, Jiaju Lin, Przemyslaw Kazienko, Jan Kocon, Jiaming Kong, Bartlomiej Koptyra, Hayden Lau, and 15 others. 2023{\natexlab{a}}.
\newblock \href {https://arxiv.org/abs/2305.13048} {Rwkv: Reinventing rnns for the transformer era}.
\newblock \emph{Preprint}, arXiv:2305.13048.

\bibitem[{Peng et~al.(2023{\natexlab{b}})Peng, Quesnelle, Fan, and Shippole}]{RoPE_3}
Bowen Peng, Jeffrey Quesnelle, Honglu Fan, and Enrico Shippole. 2023{\natexlab{b}}.
\newblock \href {https://arxiv.org/abs/2309.00071} {Yarn: Efficient context window extension of large language models}.
\newblock \emph{Preprint}, arXiv:2309.00071.

\bibitem[{Press et~al.(2022)Press, Smith, and Lewis}]{alibi}
Ofir Press, Noah Smith, and Mike Lewis. 2022.
\newblock \href {https://openreview.net/forum?id=R8sQPpGCv0} {Train short, test long: Attention with linear biases enables input length extrapolation}.
\newblock In \emph{International Conference on Learning Representations}.

\bibitem[{Shaham et~al.(2023)Shaham, Ivgi, Efrat, Berant, and Levy}]{zeroscrolls}
Uri Shaham, Maor Ivgi, Avia Efrat, Jonathan Berant, and Omer Levy. 2023.
\newblock \href {https://arxiv.org/abs/2305.14196} {Zeroscrolls: A zero-shot benchmark for long text understanding}.
\newblock \emph{arXiv preprint arXiv:2305.14196}.

\bibitem[{Song et~al.(2024)Song, Zheng, and Luo}]{countingstar}
Mingyang Song, Mao Zheng, and Xuan Luo. 2024.
\newblock \href {https://arxiv.org/abs/2403.11802} {Counting-stars: A multi-evidence, position-aware, and scalable benchmark for evaluating long-context large language models}.
\newblock \emph{arXiv preprint arXiv:2403.11802}.

\bibitem[{Su et~al.(2024)Su, Ahmed, Lu, Pan, Bo, and Liu}]{RoPE}
Jianlin Su, Murtadha Ahmed, Yu~Lu, Shengfeng Pan, Wen Bo, and Yunfeng Liu. 2024.
\newblock \href {https://doi.org/10.1016/j.neucom.2023.127063} {{{RoFormer}}: {{Enhanced}} transformer with {{Rotary Position Embedding}}}.
\newblock \emph{Neurocomputing}, 568:127063.

\bibitem[{Team(2024)}]{qwen2.5}
Qwen Team. 2024.
\newblock \href {https://qwenlm.github.io/blog/qwen2.5/} {Qwen2.5: A party of foundation models}.

\bibitem[{Wu et~al.(2022)Wu, Lan, Qian, Gu, Geramifard, and Yu}]{memformer}
Qingyang Wu, Zhenzhong Lan, Kun Qian, Jing Gu, Alborz Geramifard, and Zhou Yu. 2022.
\newblock \href {https://arxiv.org/abs/2010.06891} {Memformer: A memory-augmented transformer for sequence modeling}.
\newblock \emph{Preprint}, arXiv:2010.06891.

\bibitem[{Xiong et~al.(2023)Xiong, Liu, Molybog, Zhang, Bhargava, Hou, Martin, Rungta, Sankararaman, Oguz, Khabsa, Fang, Mehdad, Narang, Malik, Fan, Bhosale, Edunov, Lewis, Wang, and Ma}]{RoPE_1}
Wenhan Xiong, Jingyu Liu, Igor Molybog, Hejia Zhang, Prajjwal Bhargava, Rui Hou, Louis Martin, Rashi Rungta, Karthik~Abinav Sankararaman, Barlas Oguz, Madian Khabsa, Han Fang, Yashar Mehdad, Sharan Narang, Kshitiz Malik, Angela Fan, Shruti Bhosale, Sergey Edunov, Mike Lewis, and 2 others. 2023.
\newblock \href {https://arxiv.org/abs/2309.16039} {Effective long-context scaling of foundation models}.
\newblock \emph{Preprint}, arXiv:2309.16039.

\bibitem[{Xu et~al.(2023)Xu, Wang, Pan, Song, Freitag, Wang, and Li}]{instructscore}
Wenda Xu, Danqing Wang, Liangming Pan, Zhenqiao Song, Markus Freitag, William~Yang Wang, and Lei Li. 2023.
\newblock \href {https://arxiv.org/abs/2305.14282} {Instructscore: Explainable text generation evaluation with finegrained feedback}.
\newblock \emph{Preprint}, arXiv:2305.14282.

\bibitem[{yuyijiong(2023)}]{LongDataCorpus}
yuyijiong. 2023.
\newblock {LongData-Corpus}.
\newblock \url{https://huggingface.co/datasets/yuyijiong/LongData-Corpus}.
\newblock Accessed: 2023-12-20.

\bibitem[{Zeng et~al.(2024)Zeng, Xu, Wang, Zhang, Yin, Zhang, Rojas, Feng, Zhao et~al.}]{GLM-4}
Aohan Zeng, Bin Xu, Bowen Wang, Chenhui Zhang, Da~Yin, Dan Zhang, Diego Rojas, Guanyu Feng, Hanlin Zhao, and 1 others. 2024.
\newblock \href {https://arxiv.org/abs/2406.12793} {Chatglm: A family of large language models from glm-130b to glm-4 all tools}.
\newblock \emph{arXiv preprint arXiv:2406.12793}.

\bibitem[{Zhang et~al.(2024{\natexlab{a}})Zhang, Liu, Xiao, Shao, Ye, and Dou}]{AB}
Peitian Zhang, Zheng Liu, Shitao Xiao, Ninglu Shao, Qiwei Ye, and Zhicheng Dou. 2024{\natexlab{a}}.
\newblock \href {https://doi.org/10.48550/arXiv.2401.03462} {Long {{Context Compression}} with {{Activation Beacon}}}.
\newblock \emph{Preprint}, arXiv:2401.03462.

\bibitem[{Zhang et~al.(2025)Zhang, Li, Wang, Qiao, Yu, Yin, and Sun}]{factguard}
Qian-Wen Zhang, Fang Li, Jie Wang, Lingfeng Qiao, Yifei Yu, Di~Yin, and Xing Sun. 2025.
\newblock \href {https://arxiv.org/abs/2504.05607} {Factguard: Leveraging multi-agent systems to generate answerable and unanswerable questions for enhanced long-context llm extraction}.
\newblock \emph{arXiv preprint arXiv:2504.05607}.

\bibitem[{Zhang et~al.(2024{\natexlab{b}})Zhang, Chen, Hu, Xu, Chen, Hao, Han, Thai, Wang, Liu, and Sun}]{infinitebench}
Xinrong Zhang, Yingfa Chen, Shengding Hu, Zihang Xu, Junhao Chen, Moo~Khai Hao, Xu~Han, Zhen~Leng Thai, Shuo Wang, Zhiyuan Liu, and Maosong Sun. 2024{\natexlab{b}}.
\newblock \href {https://arxiv.org/abs/2402.13718} {$\infty$bench: Extending long context evaluation beyond 100k tokens}.
\newblock \emph{arXiv preprint arXiv:2402.13718}.

\bibitem[{Zhang et~al.(2021)Zhang, Ren, and de~Rijke}]{zhang-etal-2021-human}
Yangjun Zhang, Pengjie Ren, and Maarten de~Rijke. 2021.
\newblock \href {https://doi.org/10.18653/v1/2021.acl-long.436} {A human-machine collaborative framework for evaluating malevolence in dialogues}.
\newblock In \emph{Proceedings of the 59th Annual Meeting of the Association for Computational Linguistics and the 11th International Joint Conference on Natural Language Processing (Volume 1: Long Papers)}, pages 5612--5623, Online. Association for Computational Linguistics.

\bibitem[{Zhu et~al.(2023)Zhu, Wang, and Wang}]{JudgeLM}
Lianghui Zhu, Xinggang Wang, and Xinlong Wang. 2023.
\newblock \href {https://arxiv.org/abs/2310.17631} {Judgelm: Fine-tuned large language models are scalable judges}.
\newblock \emph{Preprint}, arXiv:2310.17631.

\end{thebibliography}

\appendix

\section{Distribution of the number of needles at different text lengths}

We present the distribution of the number of needles at different text lengths of the test set in Table~\ref{tab:dist_of_needles}. We not only list the distribution of the number of needles in the entire test set, but also provide the distribution of needles across different text length intervals. Both the overall distribution and the grouped distribution maintain consistency.

\begin{table}[!ht]
    \centering
    \resizebox{\columnwidth}{!}{
    \begin{tabular}{lllllll}
    \hline
        \# needles & \# QA & Proportion & 8k-16k & 16k-32k & 32k-64k & 64k-128k \\ \hline
        3-5 & 619 & 30.95\% & 79 (36\%) & 195 (32\%) & 187 (30\%) & 158 (28\%) \\ 
        6-10 & 1,028 & 51.40\% & 110 (50\%) & 309 (50\%) & 304 (50\%) & 305 (54\%) \\ 
        >10 & 353 & 17.65\% & 31 (14\%) & 106 (17\%) & 119 (20\%) & 97 (18\%) \\ 
        Total & 2,000 & 100\% & 220 & 610 & 610 & 560 \\ \hline
    \end{tabular}}
    \caption{Distribution of the number of needles.}
    \label{tab:dist_of_needles}
\end{table}

\section{Detailed Performance of Evaluation Model}

Table~\ref{tab:em_test_detail} presents the detailed performance of our evaluation model. Two critical points should be noted:
\begin{itemize}
    \item For the GT (Ground Truth) answer group, when the question doesn't require ordered output, all candidate answers should be judged as "correct". When ordered output is required, "w/ shuffle" answers should be judged as "incorrect" and "w/o shuffle" answers should be judged as "correct". Moreover, All non-GT group answers should be judged as "incorrect".
    \item Only temporal-order needles require grouping based on question requirements. Logical-order needles must always be output sequentially, thus requiring no question-based grouping.
\end{itemize}

\begin{table*}[tbhp]
  \centering
  \resizebox{\textwidth}{!}{
  \begin{tabular}{l|l|cc|cc|cc|cc}
  \toprule
  \multicolumn{10}{c}{Claude-3.5} \\ \hline
  \multirow{3}{*}{Needle types} & \multirow{3}{*}{Question} & \multicolumn{8}{c}{Answer groups} \\ \cline{3-10}
  & & \multicolumn{2}{c|}{GT} & \multicolumn{2}{c|}{Missing} & \multicolumn{2}{c|}{Redundant} & \multicolumn{2}{c}{Incorrect} \\ \cline{3-10}
  & & w/o shuffle & w/ shuffle & w/o shuffle & w/ shuffle & w/o shuffle & w/ shuffle & w/o shuffle & w/ shuffle \\ \hline
  \multirow{2}{*}{Temporal order}  & w/ order & \textcolor{red}{1} / 74         & 58 / \textcolor{red}{8}        & 64 / \textcolor{red}{7}        & 73 / \textcolor{red}{2}        & 56 / \textcolor{red}{25}       & 74 / \textcolor{red}{5}        & 59 / \textcolor{red}{2}        & 70 / \textcolor{red}{3} \\ \cline{2-10}
  & w/o order & 0 / 48        & \textcolor{red}{27} / 25       & 42 / \textcolor{red}{4}        & 41 / \textcolor{red}{2}        & 36 / \textcolor{red}{19}       & 40 / \textcolor{red}{10}       & 42 / \textcolor{red}{2}        & 47 / \textcolor{red}{4}  \\ \hline
  Logical order & w/ order & \textcolor{red}{12} / 107       & 90 / \textcolor{red}{20}       & 107 / \textcolor{red}{7}       & 125 / \textcolor{red}{10}      & 113 / \textcolor{red}{17}      & 114 / \textcolor{red}{22}      & 92 / \textcolor{red}{15}       & 110 / \textcolor{red}{29}   \\ 
  \toprule

  \multicolumn{10}{c}{GPT-4o} \\ \hline
  \multirow{3}{*}{Needle types} & \multirow{3}{*}{Question} & \multicolumn{8}{c}{Answer groups} \\ \cline{3-10}
  & & \multicolumn{2}{c|}{GT} & \multicolumn{2}{c|}{Missing} & \multicolumn{2}{c|}{Redundant} & \multicolumn{2}{c}{Incorrect} \\ \cline{3-10}
  & & w/o shuffle & w/ shuffle & w/o shuffle & w/ shuffle & w/o shuffle & w/ shuffle & w/o shuffle & w/ shuffle \\ \hline
  \multirow{2}{*}{Temporal order}  & w/ order & 0 / 75        & 59 / \textcolor{red}{7}        & 67 / \textcolor{red}{4}        & 75 / 0        & 68 / \textcolor{red}{13}       & 72 / \textcolor{red}{7}        & 61 / 0        & 72 / \textcolor{red}{1}     \\ \cline{2-10}
  & w/o order & 0 / 48        & \textcolor{red}{15} / 37        & 45 / \textcolor{red}{1}        & 42 / \textcolor{red}{1}        & 47 / \textcolor{red}{8}        & 39 / \textcolor{red}{11}       & 44 / 0        & 46 / \textcolor{red}{5}    \\ \hline
  Logical order & w/ order & 0 / 119       & 108 / \textcolor{red}{2}       & 114 / 0       & 135 / 0       & 129 / \textcolor{red}{1}       & 135 / \textcolor{red}{1}       & 107 / 0       & 139 / 0     \\ 
  \toprule

  \multicolumn{10}{c}{Our Evaluation Model} \\ \hline
  \multirow{3}{*}{Needle types} & \multirow{3}{*}{Question} & \multicolumn{8}{c}{Answer groups} \\ \cline{3-10}
  & & \multicolumn{2}{c|}{GT} & \multicolumn{2}{c|}{Missing} & \multicolumn{2}{c|}{Redundant} & \multicolumn{2}{c}{Incorrect} \\ \cline{3-10}
  & & w/o shuffle & w/ shuffle & w/o shuffle & w/ shuffle & w/o shuffle & w/ shuffle & w/o shuffle & w/ shuffle \\ \hline
  \multirow{2}{*}{Temporal order}  & w/ order & 0 / 75        & 64 / \textcolor{red}{2}        & 71 / 0        & 75 / 0        & 81 / 0        & 79 / 0        & 61 / 0        & 73 / 0  \\ \cline{2-10}
  & w/o order & 0 / 48        &  \textcolor{red}{8} / 44        & 46 / 0        & 43 / 0        & 55 / 0        & 50 / 0        & 44 / 0        & 51 / 0    \\ \hline
  Logical order & w/ order & 0 / 119       & 110 / 0       & 114 / 0       & 135 / 0       & 130 / 0       & 136 / 0       & 107 / 0       & 139 / 0    \\ 
  \bottomrule
  \end{tabular}}
  \caption{Detailed performance of evaluation model compared with Claude-3.5 and GPT-4o on synthetic test data. The `w/ order' vs. `w/o order' indicates whether the question requires LLMs to output results in temporal order. The `w/ shuffle' vs. `w/o shuffle' condition determines whether the answer candidates were shuffled before evaluation. The `A/B' ratio reflects [incorrect/correct] judgments obtained by LLMs within the test subset. The \textcolor{red}{red font} indicates the number of misjudged samples that deviate from expected results.}
  \label{tab:em_test_detail}
\end{table*}

\section{Failure modes analysis}

The evaluation model we trained not only determines whether an answer is correct, but also provides the rationale behind its judgment. This allows us to analyze the primary sources of failure modes based on these diagnostic explanations. Tabel~\ref{tab:failure_modes_analysis} shows our findings.

\begin{table*}[!ht]
    \centering
    \resizebox{\textwidth}{!}{
    \begin{tabular}{lcccccc}
    \hline
        Model & Missing items & Redundant items & Incorrect items & Wrong order & Others & No. incorrect samples \\ \hline
        Claude-3.5 & 851 (77.58\%) & 177 (16.13\%) & 222 (20.24\%) & 276 (25.16\%) & 149 (13.58\%) & 1097 \\ 
        GPT-4o & 956 (66.71\%) & 338 (23.59\%) & 249 (17.38\%) & 189 (13.19\%) & 936 (65.32\%) & 1433 \\
        GPT-4o-mini & 1111 (79.70\%) & 166 (11.91\%) & 269 (19.30\%) & 318 (22.81\%) & 135 (9.68\%) & 1394 \\ 
        Gemini-1.5 & 471 (64.52\%) & 99 (13.56\%) & 112 (15.34\%) & 171 (23.42\%) & 118 (16.16\%) & 730 \\
        LLaMA-3.3 & 880 (73.03\%) & 112 (9.29\%) & 262 (21.74\%) & 371 (30.79\%) & 29 (2.41\%) & 1205 \\ 
        Qwen-2.5-72B & 724 (72.47\%) & 128 (12.81\%) & 253 (25.33\%) & 283 (28.33\%) & 10 (1.00\%) & 999 \\ 
        Qwen-2.5-32B & 919 (75.89\%) & 136 (11.23\%) & 303 (25.02\%) & 391 (32.29\%) & 35 (2.89\%) & 1211 \\ 
        Qwen-2.5-7B & 994 (76.23\%) & 192 (14.72\%) & 331 (25.38\%) & 395 (30.29\%) & 28 (2.15\%) & 1304 \\ 
        Qwen-3-32B & 814 (73.60\%) & 193 (17.45\%) & 289 (26.13\%) & 322 (29.11\%) & 28 (2.53\%) & 1106 \\ 
        Qwen-3-14B & 912 (73.14\%) & 253 (20.29\%) & 306 (24.54\%) & 402 (32.24\%) & 35 (2.81\%) & 1247 \\ 
        Qwen-3-8B & 883 (63.85\%) & 401 (28.99\%) & 536 (38.76\%) & 433 (31.31\%) & 60 (4.34\%) & 1383 \\ \hline
    \end{tabular}
    }
    \caption{Failure modes analysis of different LLMs. All incorrectly evaluated samples were categorized by failure type, and the percentage for each category is provided.}
  \label{tab:failure_modes_analysis}
\end{table*}

We analyzed the proportion of different types of failure modes in the erroneous samples of each LLM based on benchmark evaluation results. Please note that the same error sample may contain more than one error pattern. Therefore, the sum of the percentages in each row of the table is greater than 1. we can find that Missing answer items is the most prevalent failure mode. "Others" generally refers to cases where the output is irrelevant to the question or contains additional content. Notably, the Qwen series of models exhibit a lower frequency of this error pattern, suggesting that they have a stronger ability to understand the question in long-context scenarios.

\section{Noise Analysis Across Groups}

\begin{figure*}[ht]
    \centering
    \begin{subfigure}[b]{0.48\textwidth}
        \includegraphics[width=\textwidth]{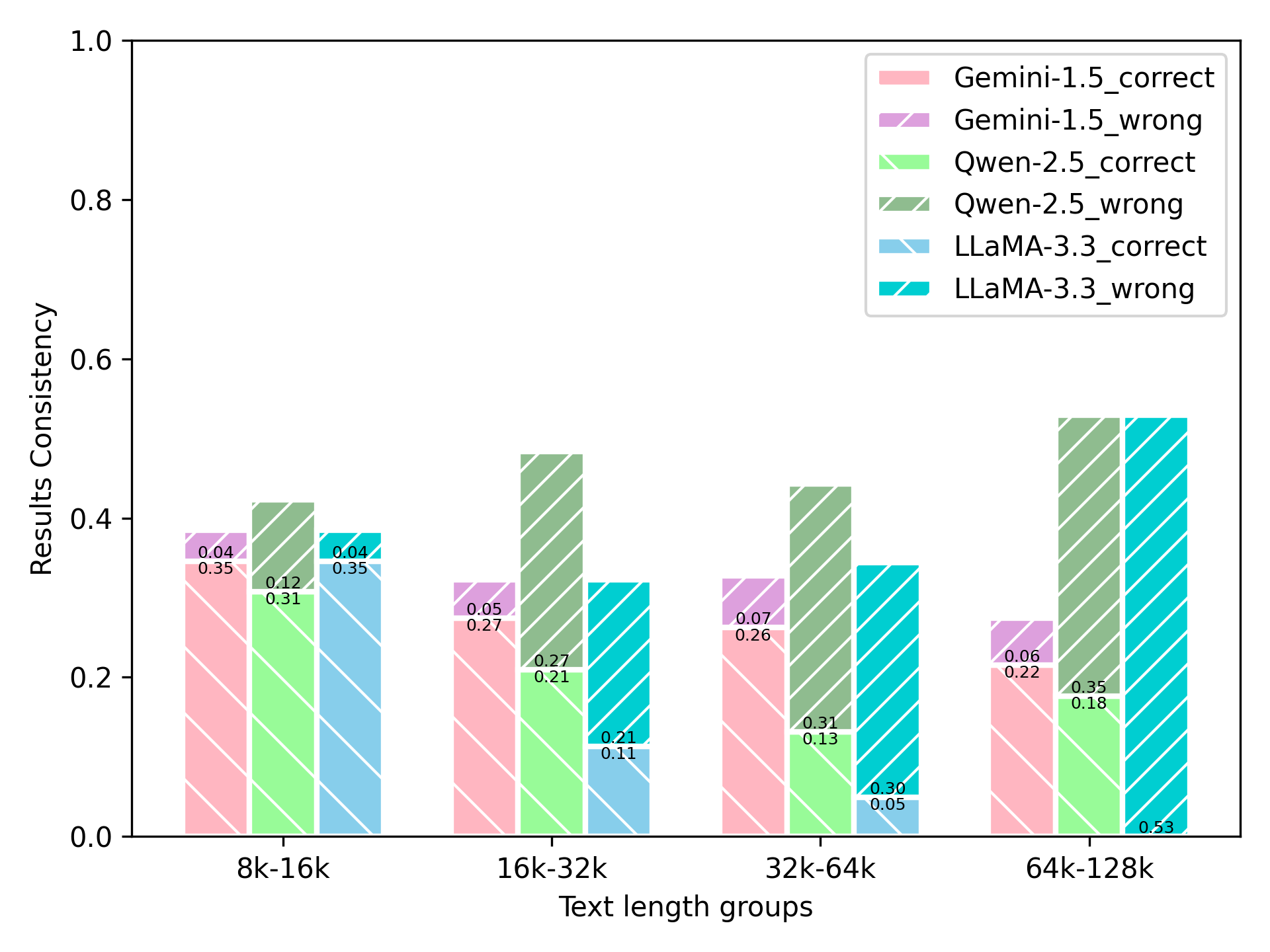}
        \caption{Results across length groups.}
        \label{fig:noise_result_1}
    \end{subfigure}
    \hfill
    \begin{subfigure}[b]{0.48\textwidth}
        \includegraphics[width=\textwidth]{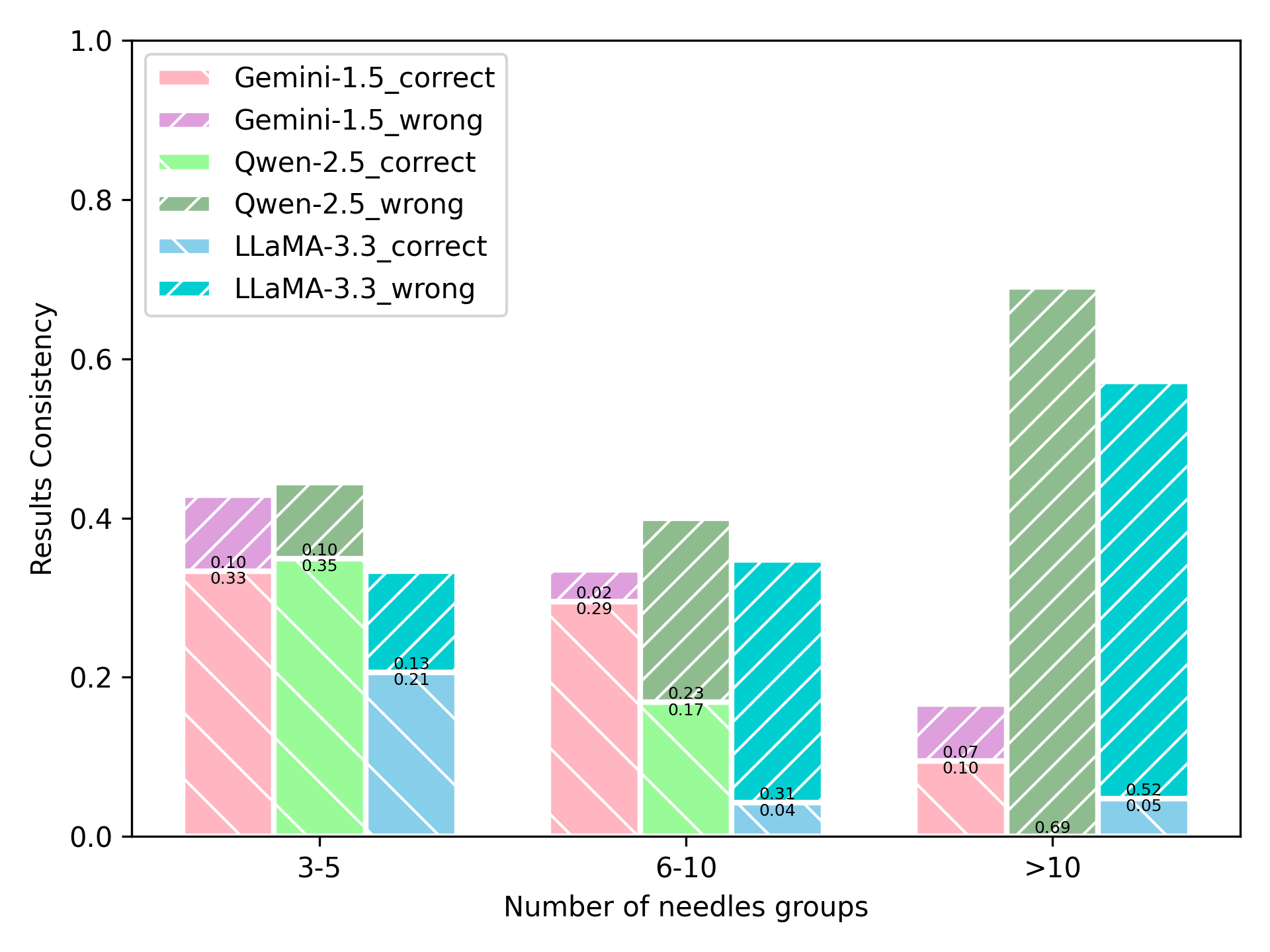}
        \caption{Results across the number of needles groups.}
        \label{fig:noise_result_2}
    \end{subfigure}
    \caption{Noise analysis results of LLMs on noise test data across different groups with correct consistency and wrong consistency.}
    \label{fig:noise}
\end{figure*}

Figure~\ref{fig:noise} presents the result consistency metric of the LLMs' responses across all noise test groups, organized by different text lengths and numbers of needles. The data shows that variations in result consistency across different text lengths are minimal, suggesting that the complexity of test samples constructed by this benchmark is largely uniform across various text lengths. However, as text length increases, there is a rise in the proportion of consistently wrong answers. This trend indicates that the task becomes more challenging with longer texts, making it more difficult to maintain model accuracy by adjusting the needles. Similarly, when the number of needles is 10 or fewer, the variation in result consistency remains small. However, when the number of needles surpasses 10, there is a marked increase in result consistency. This rise is primarily due to the higher task difficulty associated with a larger number of needles, leading to a corresponding increase in consistently wrong answers, which aligns with expectations.

\section{Noise Example}

\begin{figure*}[th]
  \includegraphics[width=\textwidth]{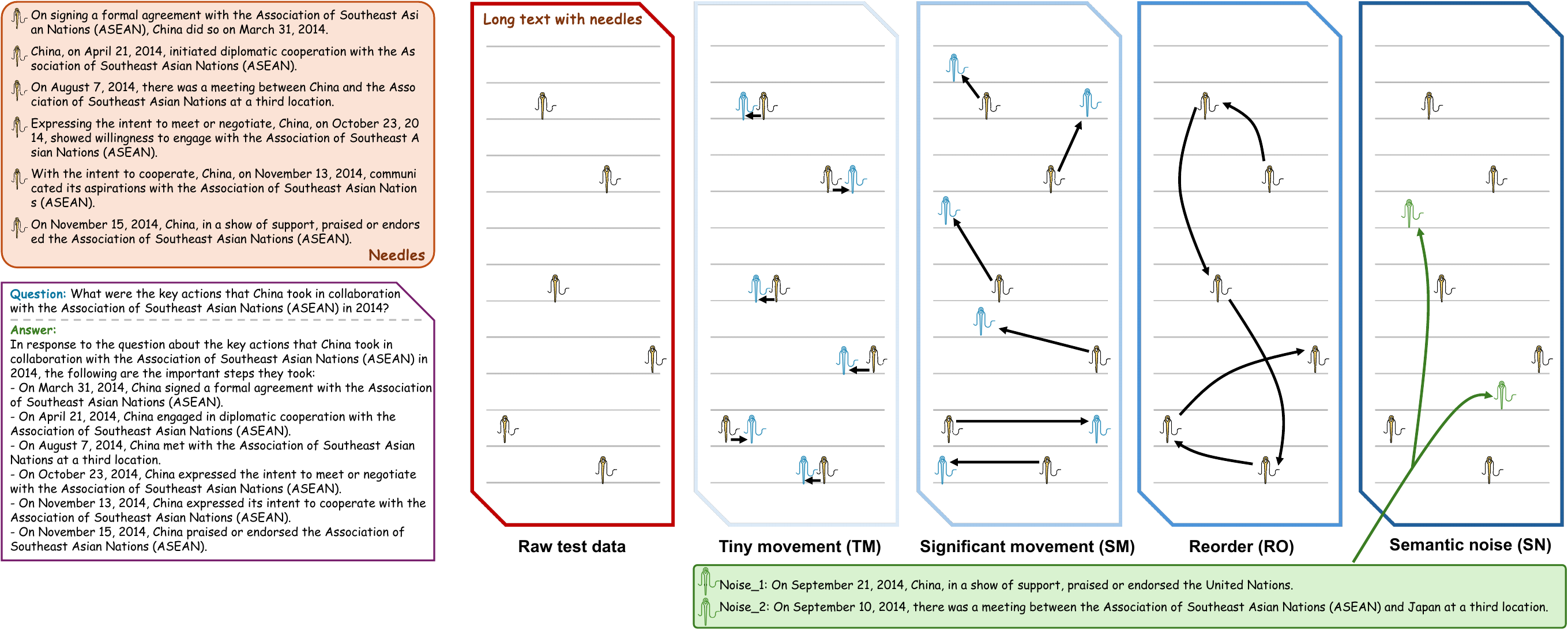}
  \caption{Four types of noise added into the raw test data.}
  \label{fig:noise_example}
\end{figure*}

Figure~\ref{fig:noise_example} provides a schematic diagram illustrating the addition of different types of noise to the raw test data.

\section{Computational resources and time requirements}

\subsection{Inference stage}

For closed-source models, the main cost comes from API calls, requiring about 97 million tokens for a full test set inference. Response time is 4-7 seconds per request, and multi-threading can improve efficiency if API bandwidth allows.

For open-source models, inference is deployed on a server with 8 H20 GPUs, with response times ranging from 3-5 seconds per request, depending on the model size. Detailed stats will be available in the appendix.

\subsection{Evaluation stage}

The evaluation model is fine-tuned on Qwen2.5-Instruct-32B and deployed on a server with 8 H20 GPUs. Response time is about 0.5 seconds per request, and a full evaluation takes around 16 minutes, faster with multi-threading.

\section{Prompts}

\subsection{Sequential-NIAH data example}

\begin{figure*}[th]
  \includegraphics[width=\textwidth]{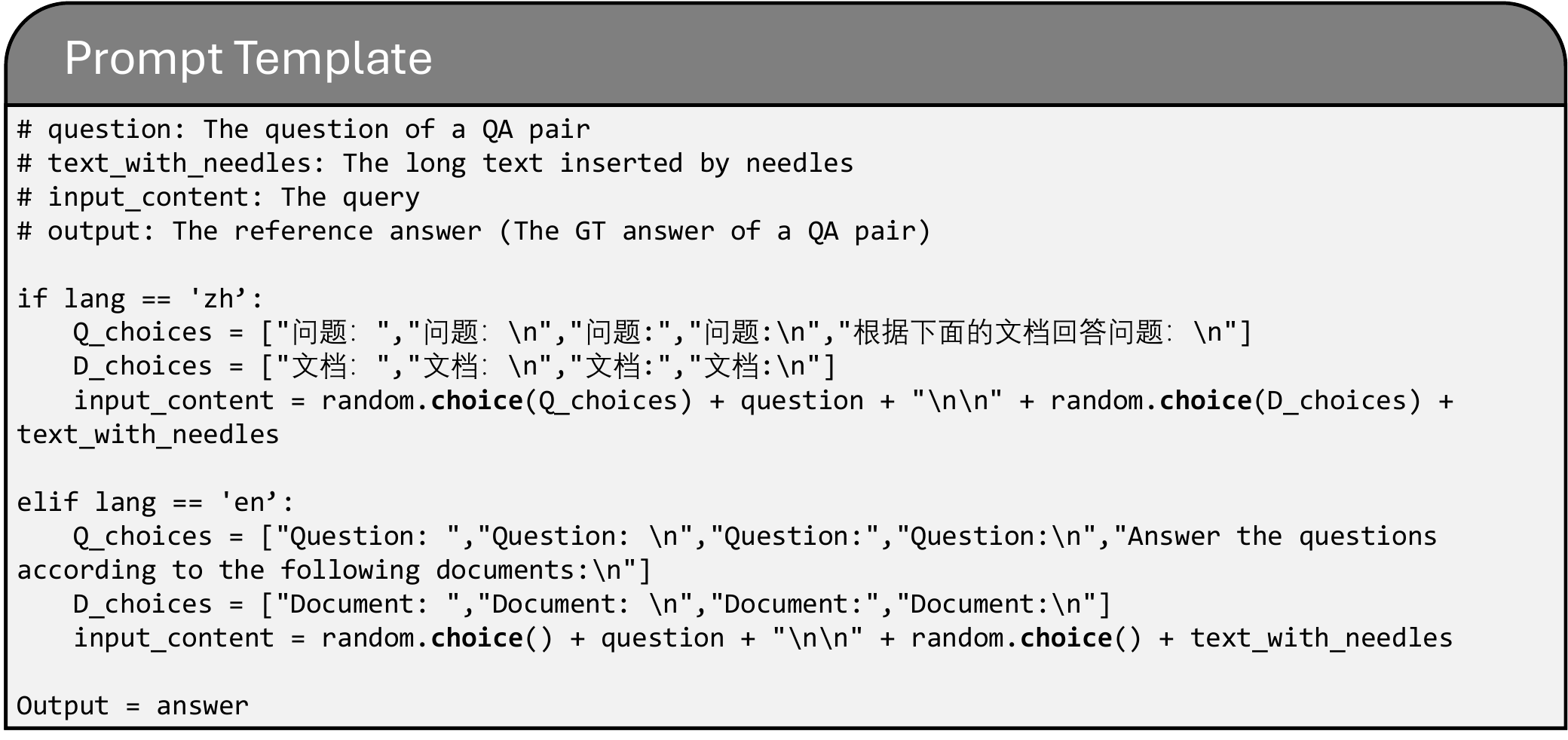}
  \caption{Prompt to build a Sequential-NIAH data.}
  \label{fig:niah_data_template}
\end{figure*}

In Figure~\ref{fig:niah_data_template}, we provide prompts for building a Sequential-NIAH data.

\subsection{Prompts for evaluation model}

\begin{figure*}[th]
  \includegraphics[width=\textwidth]{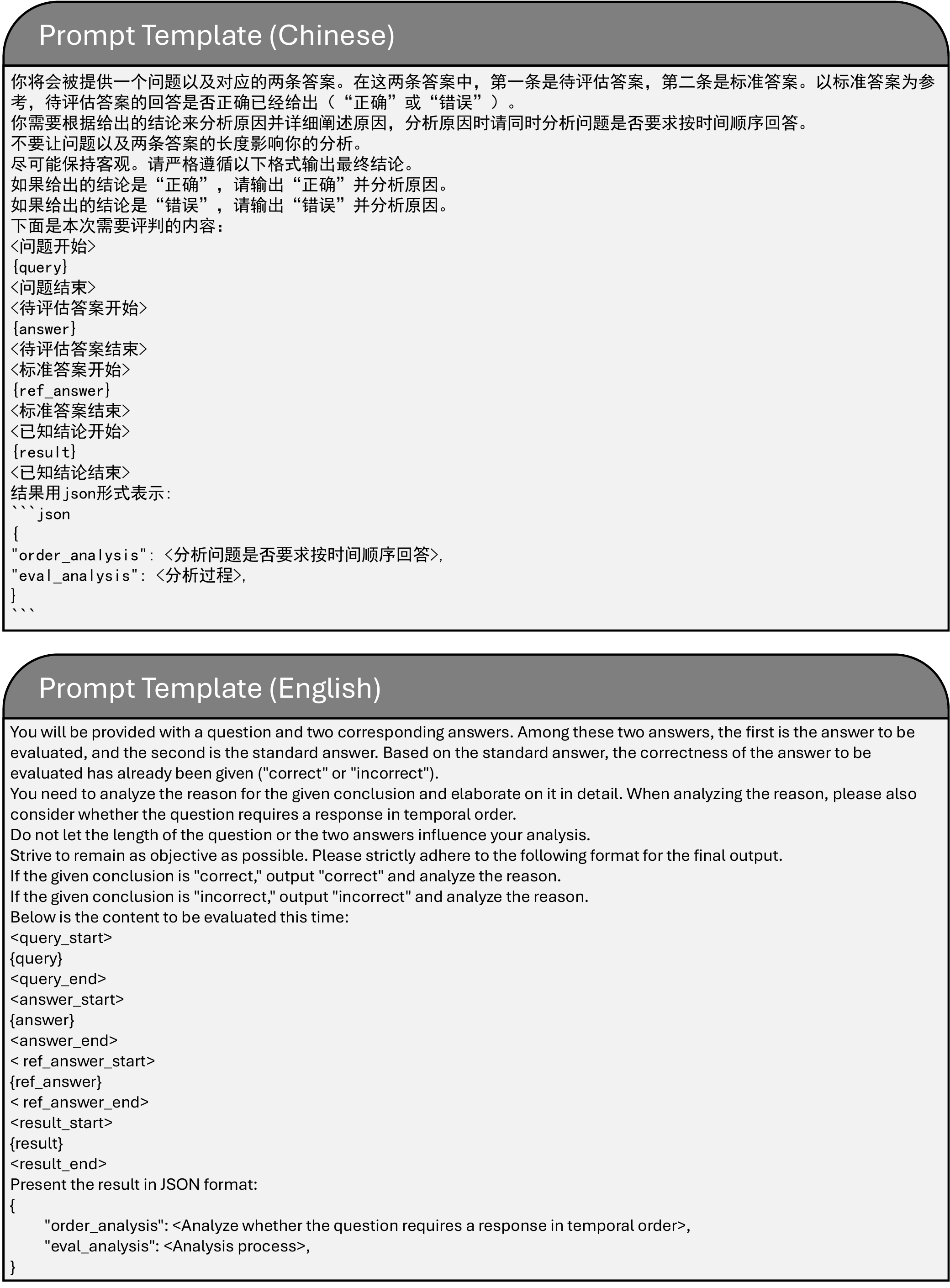}
  \caption{Prompt for reason analysis to build the training data of evaluation model.}
  \label{fig:prompt_for_reason}
\end{figure*}

In Figure~\ref{fig:prompt_for_reason}, we provide the prompt to obtain the $R_i$ for building the training data of our evaluation model, which is the prompt for reason analysis in Figure~\ref{fig:EM}.

\begin{figure*}[th]
  \includegraphics[width=\textwidth]{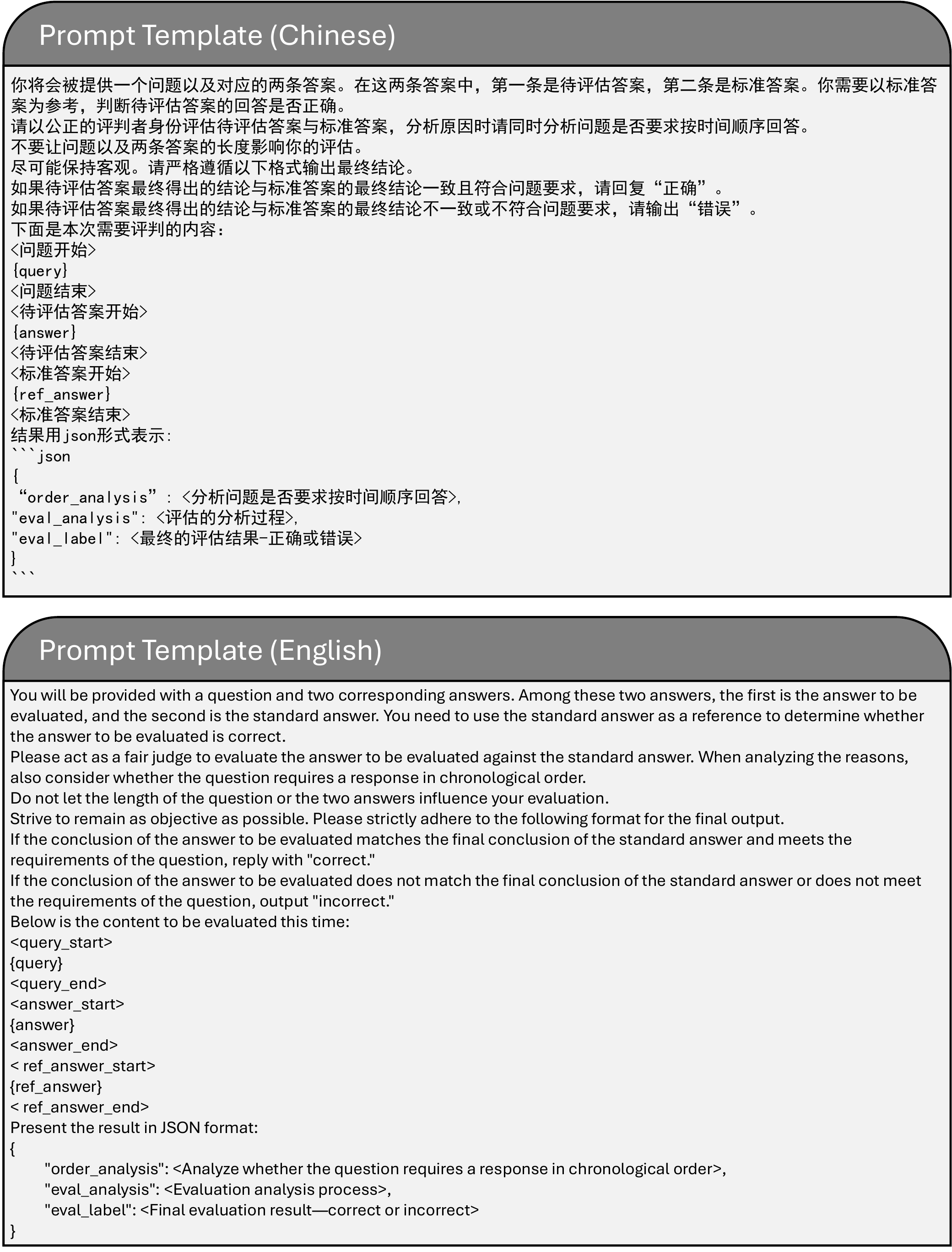}
  \caption{Prompt for evaluating the response of LLMs for temporal-order needles extraction.}
  \label{fig:prompt_for_em_train_1}
\end{figure*}

In Figure~\ref{fig:prompt_for_em_train_1}, we provide the prompt for evaluating the response of LLMs for temporal-order needles extraction.

\begin{figure*}[th]
  \includegraphics[width=\textwidth]{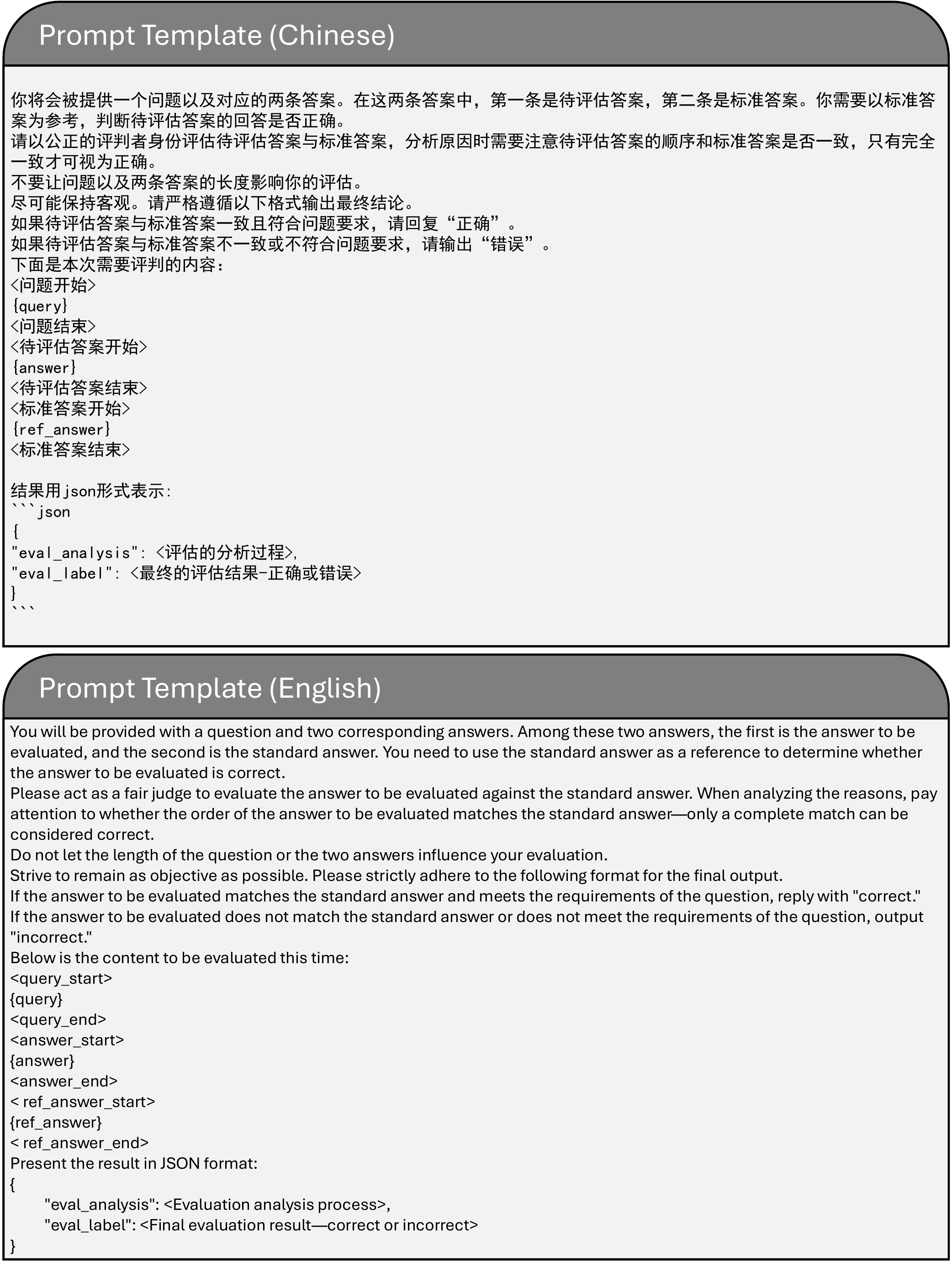}
  \caption{Prompt for evaluating the response of LLMs for logical-order needles extraction.}
  \label{fig:prompt_for_em_train_2}
\end{figure*}

In Figure~\ref{fig:prompt_for_em_train_2}, we provide the prompt for evaluating the response of LLMs for logical-order needles extraction.

\subsection{Prompts for synthetic-temporal order needles}

\begin{figure*}[th]
  \includegraphics[width=\textwidth]{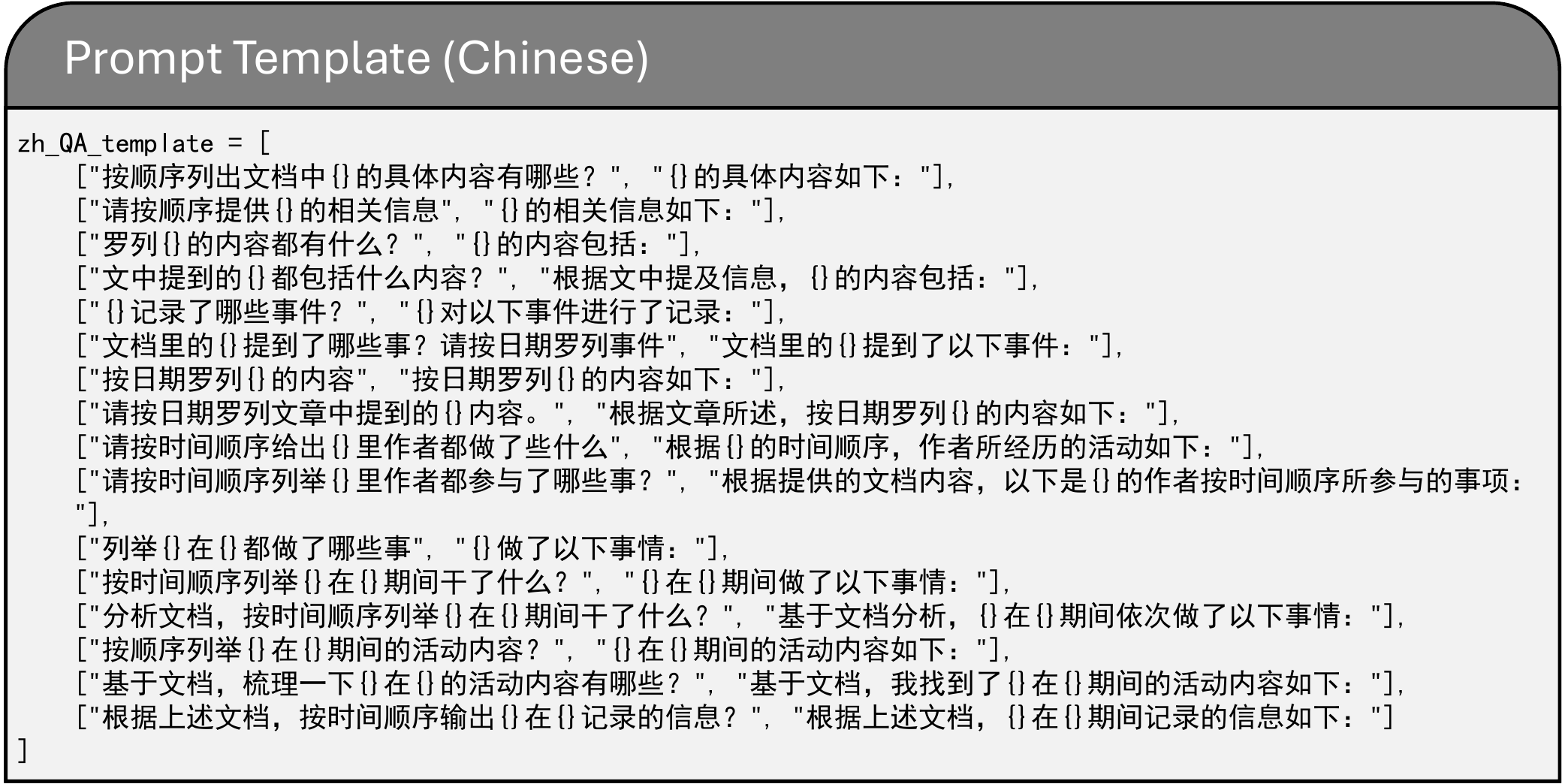}
  \caption{Prompt for QA template of Chinese synthetic-temporal order needles.}
  \label{fig:prompt_for_syn_zh}
\end{figure*}

In Figure~\ref{fig:prompt_for_syn_zh}, we provide the prompt of QA template of Chinese synthetic-temporal order needles.

\begin{figure*}[th]
  \includegraphics[width=\textwidth]{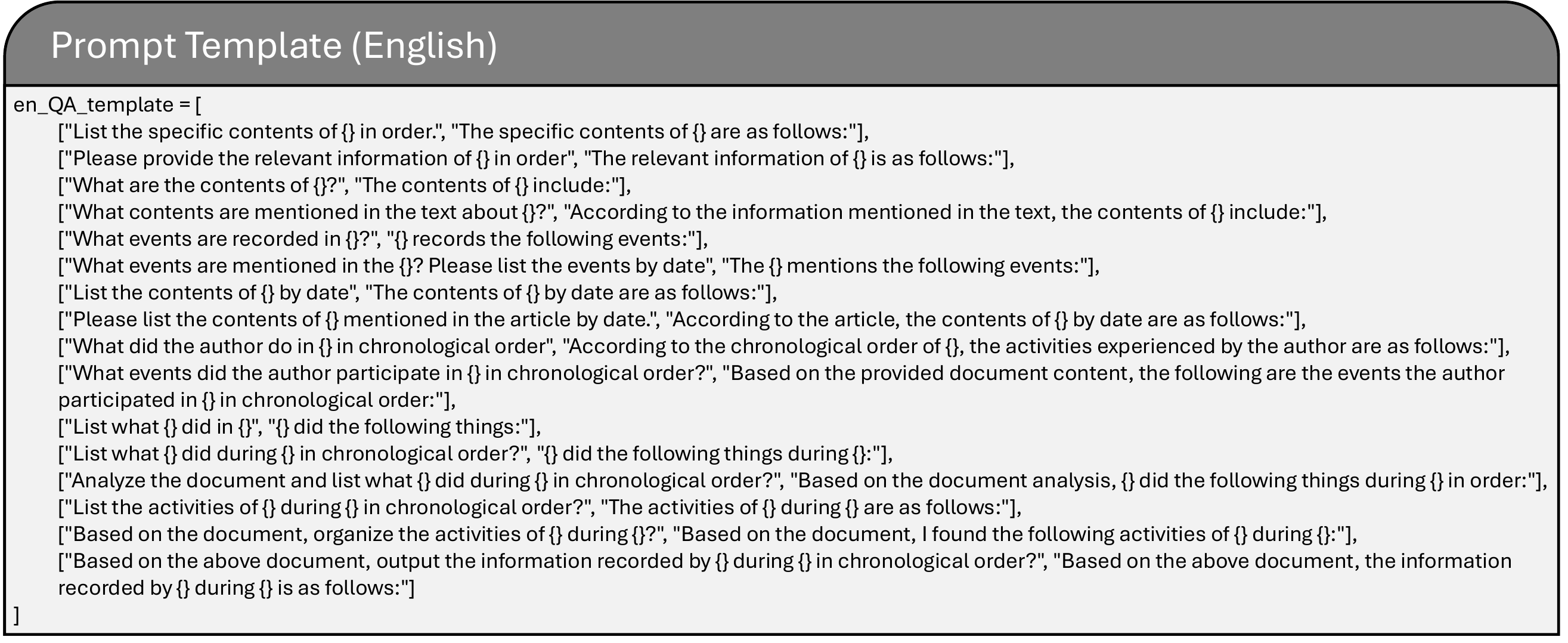}
  \caption{Prompt for QA template of English synthetic-temporal order needles.}
  \label{fig:prompt_for_syn_en}
\end{figure*}

In Figure~\ref{fig:prompt_for_syn_en}, we provide the prompt of QA template of English synthetic-temporal order needles.

\subsection{Prompts for Real-temporal order needles}

\begin{figure*}[th]
  \includegraphics[width=\textwidth]{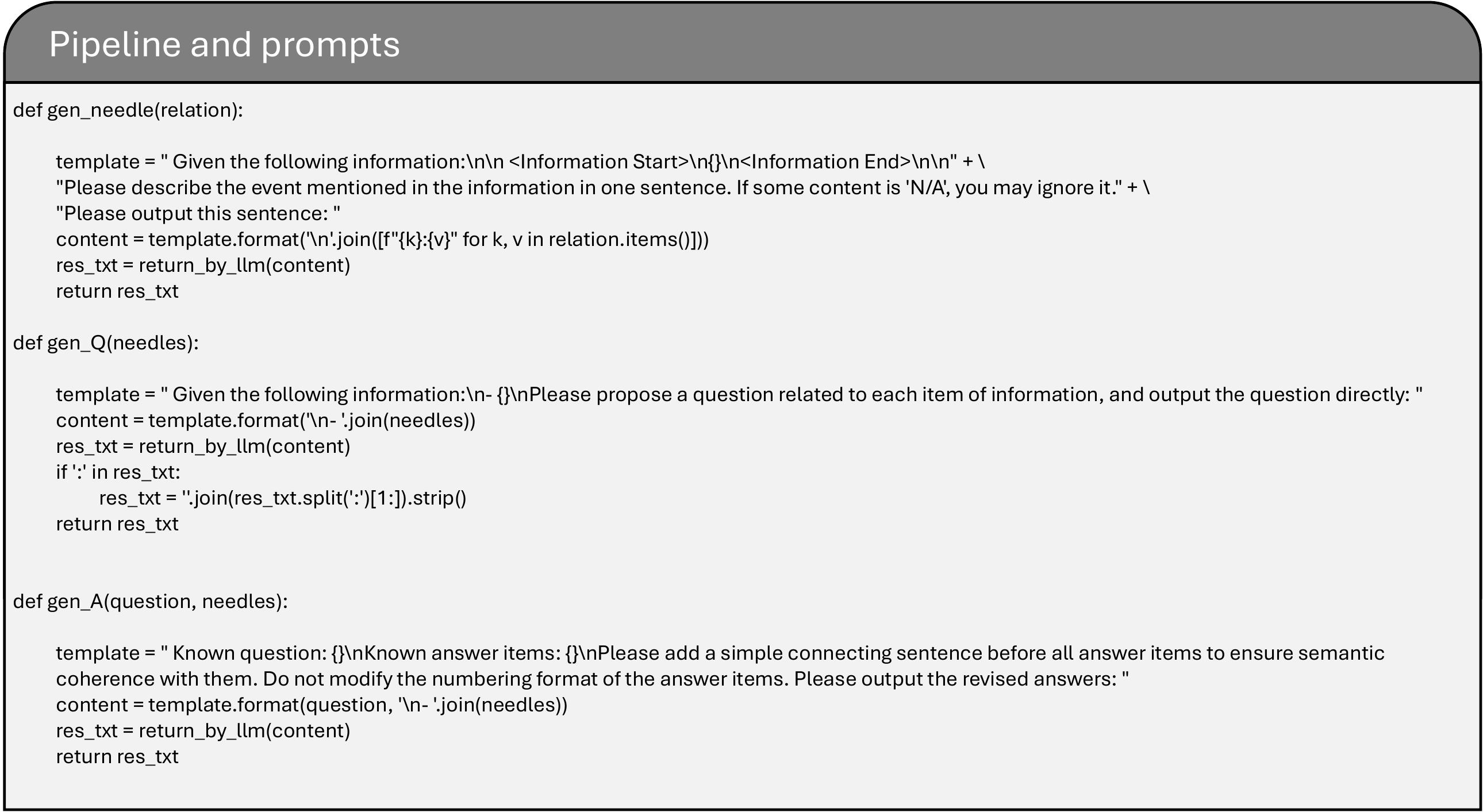}
  \caption{Pipeline and prompts for needles and QA generation of real-temporal order needles.}
  \label{fig:prompt_for_real}
\end{figure*}

In Figure~\ref{fig:prompt_for_real}, we provide the prompt of QA template of English synthetic-temporal order needles.

\subsection{Prompts for Real-logical order needles}

\begin{figure*}[th]
  \includegraphics[width=\textwidth]{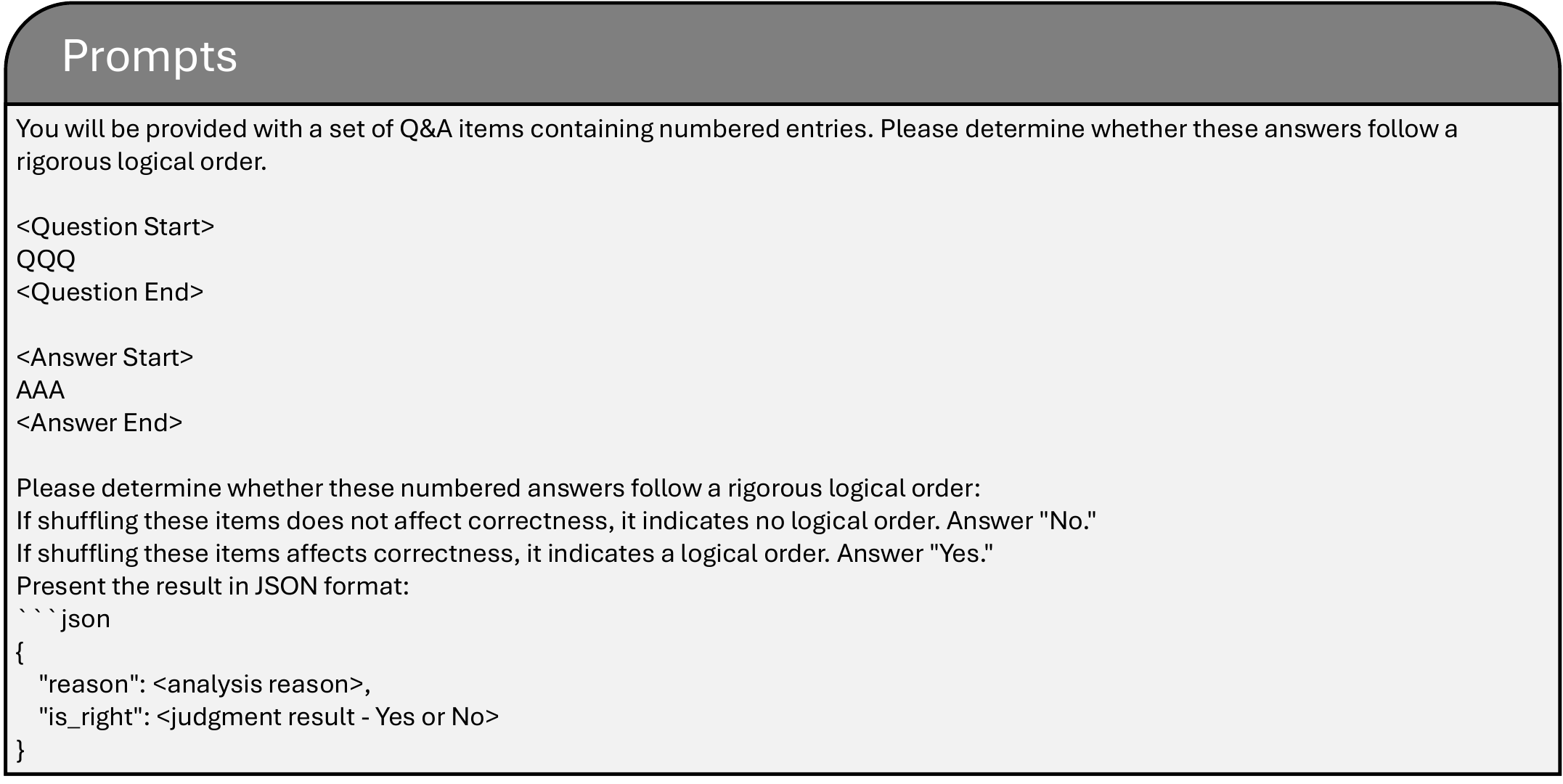}
  \caption{Prompt for QA screening from QA pool.}
  \label{fig:prompt_for_qa_screen}
\end{figure*}

In Figure~\ref{fig:prompt_for_qa_screen}, we provide the prompt of QA screening from QA pool.

\begin{figure*}[th]
  \includegraphics[width=\textwidth]{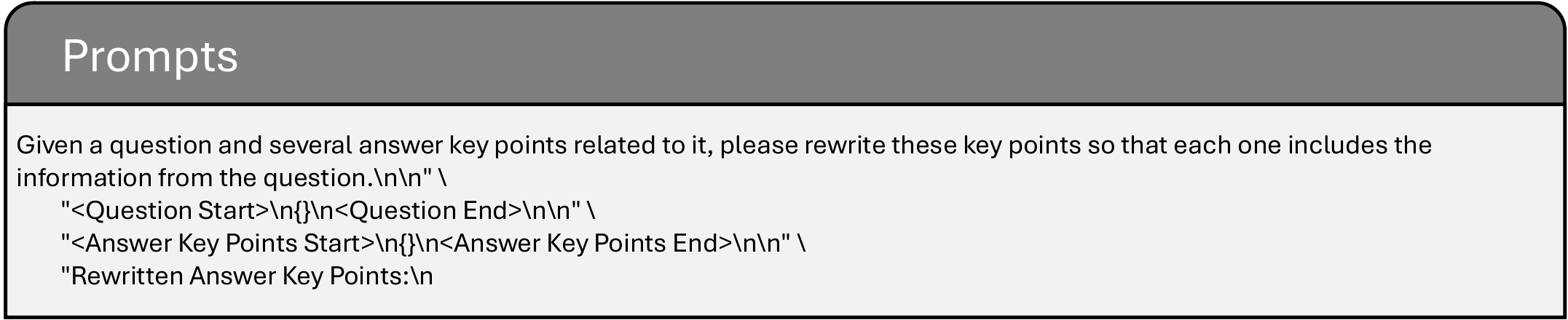}
  \caption{Prompt for needles generation based on answer items.}
  \label{fig:prompt_for_item_modify}
\end{figure*}

In Figure~\ref{fig:prompt_for_item_modify}, we provide the prompt of needles generation based on answer items.

\end{document}